\documentclass{article}

 \usepackage[preprint]{neurips_2026}

\usepackage[utf8]{inputenc} 
\usepackage[T1]{fontenc}    
\usepackage{hyperref}       
\usepackage{url}            
\usepackage{booktabs}       
\usepackage{multirow}
\usepackage{amsfonts}       
\usepackage{nicefrac}       
\usepackage{microtype}      
\usepackage{xcolor}         
\usepackage{graphicx}
\usepackage{subcaption}
\usepackage{amsmath}
\usepackage{cleveref}
\usepackage{wrapfig}
\usepackage{enumitem}

\DeclareMathOperator*{\argmax}{arg\,max}

\title{Gen4U: Unifying Video Generation and Understanding via Diffusion}

\author{%
  Michael King\thanks{equal contribution}\\
  Google DeepMind \\
  \And Aravindh Mahendran\footnotemark[1] \\
  Google DeepMind \\
  \And Matthew Koichi Grimes \\
  Google DeepMind \\
  \And Fedor Kitashov \\
  Google DeepMind \\
  \And Adham Elarabawy \\
  Google DeepMind \\
  \And Pedro Velez \\
  Google DeepMind \\
  \And Maks Ovsjanikov \\
  Google DeepMind
  \And Viorica P\u atr\u aucean\thanks{corresponding author \texttt{viorica@google.com}} \\
  Google DeepMind
}

\begin{document}

\maketitle

\begin{abstract}
Prior work suggests that diffusion representations capture low-level geometry but struggle with high-level semantics. We demonstrate that state-of-the-art video diffusion models overcome this limitation. By systematically probing their intermediate activations using recent mutual-kNN alignment metrics, we reveal a highly structured latent space where visual representations evolve across both network depth and noise levels. We show that while moderate noise levels yield linearly separable global semantics, fine-grained  details persist at lower noise levels but become spatially scattered, requiring attention mechanisms to decode. Building on these insights, we introduce Gen4U (Generation for Understanding), a framework that repurposes these generative representations with a single forward pass. Our experiments establish that frozen, large-scale video diffusion models function as highly competitive video encoders across a wide spectrum of tasks, spanning semantic and non-semantic objectives (video classification, depth estimation, camera pose estimation, image and video captioning). Bypassing fine-tuning, Gen4U unifies the generation and understanding paradigms, achieving strong perception performance while fully preserving the model's ability to generate high-quality video.  
\end{abstract}

\section{Introduction}
\label{sec:intro}

Current paradigms in visual representation learning struggle to reconcile geometry with semantics. Contrastive methods and large Vision-Language Models (VLMs)~\citep{clip,beyer2024paligemma,comanici2025gemini25pushingfrontier} operate primarily in language space, yielding strong semantics but neglecting precise spatio-temporal details. Conversely, pixel-level reconstruction approaches like masked auto-encoding (MAE)~\citep{tong2022videomae,wang2023videomae,carreira2025scaling4drepresentations,zhang2026efficientlyreconstructingdynamicscenes} excel at preserving local geometry and motion but struggle to generalise to broad semantic tasks. A single, task-agnostic foundation video model that supports both types of capabilities remains an open problem.

Inspired by the predictive coding theory of the brain, which posits that prediction is key to cognition~\citep{Rao1999PredictiveCI,clark}, we investigate how to build general-purpose video models based on generative video diffusion transformers. We argue that generating temporally and geometrically consistent video requires the model to implicitly learn the underlying mechanics of the physical world, such as motion, object permanence, and part-object relationships. However, recent studies probing frozen video diffusion representations concluded that they capture low-level geometry but struggle with high-level semantics, casting doubt on their eligibility as general-purpose perception encoders~\citep{Velez_2025_ICCV,ovsjanikov2026dynamic}.

In this work, we demonstrate that state-of-the-art video diffusion models do, in fact, overcome this limitation as they become more capable at generating physically and temporally consistent videos. By systematically investigating the intermediate activations of both proprietary (Veo3~\citep{deepmind2025veo3}) and open-weight (Wan2.2-T2V-A14B~\citep{wan2025}) models, we map the hierarchy of their latent spaces using recent zero-shot mutual k-NN alignment metrics~\citep{pmlr-v235-huh24a,ovsjanikov2026dynamic} and we identify optimal spots for extracting linearly separable global semantics and more entangled, spatially-distributed features that require attention-based pooling to extract effectively.  

Based on these observations, we introduce \textit{Gen4U} (Generation for Understanding), a framework that extracts and repurposes generative representations for a diverse set of downstream visual tasks. Because Gen4U identifies the optimal block and noise level for feature extraction, it requires only a single forward pass through the diffusion backbone. This makes it as computationally efficient as a standard discriminative visual encoder, e.g.~\citep{carreira2025scaling4drepresentations,internvideo2}, circumventing the high inference costs typically associated with iterative denoising. By pairing these frozen representations with lightweight decoders, Gen4U achieves strong perception performance without altering the pre-trained generative weights. Our extensive evaluations demonstrate that frozen video diffusion models can function as general-purpose video encoders. 

To summarise, our main contributions are: 

\begin{enumerate}[label=\alph*.]
    \item \textit{Latent space analysis}: We map the evolution of diffusion features across depth and noise levels using recent alignment metrics, identifying the optimal points to extract representations for understanding.
    \item \textit{Semantics and temporal dynamics}: We show that Gen4U captures rich, high-level understanding, achieving state-of-the-art results on zero-shot alignment and video classification on Something-Something V2 dataset~\citep{goyal2017something}, alongside satisfactory results on image and video captioning.
    \item \textit{Geometry understanding}: We demonstrate that these same frozen representations excel at lower level geometry awareness, achieving strong performance on monocular depth estimation and camera pose estimation.
\end{enumerate}

\section{Related work}
\label{sec:related}

The most common paradigms for learning visual representations are variations of masked autoencoding (MAE)~\citep{He2021MaskedAA,tong2022videomae,wang2023videomae}, scaled in 4DS~\citep{carreira2025scaling4drepresentations} and D4RT~\citep{zhang2026efficientlyreconstructingdynamicscenes}, and contrastive learning, with or without negative pairs (SigLip~\citep{zhai2023sigmoid}, SimCLR \citep{simclr}, CLIP \citep{clip}, BYOL~\citep{byol}, BRAVE~\citep{brave}, DINO~\citep{dino} to name a few). These methods generally strike a trade-off between low level scene understanding and high level semantics, failing to support both. 
Even hybrid approaches~\citep{videoprism,internvideo2,Papalampidi2023ASR} that combine masked auto-encoding (MAE) with contrastive learning can still accurately span only one category of tasks. More recently, the V-JEPA line of works~\citep{bardes2023vjepa,assran2025vjepa2selfsupervisedvideo,murlabadia2026vjepa21unlockingdense} focus on predicting the future in latent space to avoid the computational cost of pixel-level prediction, and show strong results on a wide range of understanding tasks. These pre-training methods lead to strong results on downstream tasks, but most of the time they require extensive fine-tuning or have fairly complex pre-training pipelines, balancing different optimisation objectives. Moreover none of these pre-training methods support high-quality video generation. 

Driven by the impressive realism and diversity of the outputs generated by diffusion models~\citep{Rombach_2022_CVPR,Gupta2023PhotorealisticVG,deepmind2025veo3,wan2025}, multiple works have started to investigate their understanding capabilities. In~\citep{Li_2023_ICCV} the authors show that a diffusion-based 
setup can perform zero-shot classification by conditioning the generative model with the possible classes and selecting the class that leads to the best reconstruction of the noise added to the input image. In~\citep{luo2023diffusion}, the authors build diffusion hyperfeatures by combining the activations from different blocks and noise levels to build robust pixel-level descriptors.  In~\citep{wiedemer2025videomodelszeroshotlearners}, the authors show impressive generation results by prompting the Veo3 model with challenging tasks that require geometry and physics understanding, hinting that Veo3 has implicitly learnt a world model. In addition, works like~\citep{ha2018worldmodels},~\citep{hafner2025mastering},~\citep{genie,parkerholder2024genie2} confirm that generative models support learning a world model for simulated environments, with representations that can be used to inform action policies.
 
The closest work to our approach is~\citep{Velez_2025_ICCV}. The authors repurpose activations extracted from variants of Walt model~\citep{Gupta2023PhotorealisticVG} adapted for image and video, for understanding tasks and conclude that diffusion models excel at lower level understanding problems like depth or camera pose estimation, but struggle with high-level semantic tasks. Similarly, in~\citep{ovsjanikov2026dynamic}, the authors measure the alignment of representations produced by different visual encoders amongst themselves and with language embeddings. Interestingly, their findings suggest that diffusion models like Walt produce representations that align with established visual encoders like DINOv2~\citep{oquab2024dinov} and VideoMAEv2~\citep{wang2023videomae}, but the alignment with text embeddings is significantly lower. In our work, we show that the recent generation of video diffusion models have strong semantic awareness, in addition to geometry and motion understanding, constituting a strong foundation for a complete perception system. We show state-of-the-art results on video classification on the challenging SSv2 dataset~\citep{goyal2017something}, compared to strong baselines from the other families of visual pre-training methods (MAE, contrastive, latent prediction). To further the investigation of semantic understanding, we set up captioning experiments, showing satisfactory performance compared to a strong SigLIP-based baseline~\citep{zhai2023sigmoid} from the PaliGemma family~\citep{beyer2024paligemma}. 

The interplay between generation and understanding started to receive more attention recently. For example, detailed point tracking conditioning leads to more consistent generation~\citep{Jeong_2025_CVPR}, or detailed language captioning produced by Gemini models lead to improved Veo generation~\citep{deepmind2025veo3}. Transfusion~\citep{zhou2025transfusion} elegantly combines generation and understanding pre-training into a single model, allowing the two objectives to support each other. However, in all these works, the main finding has been that advanced understanding models lead to significant improvement in generative models. The other direction, how can generation help understanding, has so far been explored mainly for low-level geometry tasks. In our work, we show that frozen diffusion generative models can power video understanding and achieve competitive performance on a wide range of visual tasks, while being able to generate high-quality videos. This unifies video \textit{generation} and \textit{understanding} paradigms. We estimate that this finding can be very impactful for the community, by encouraging more research into understanding the representation power of diffusion models. In addition, the implications from a practical point of view could be substantial, as it could lead to one visual model shared across generation and understanding.  
\section{The structure of the generative manifold}
\label{sec:method}

The iterative diffusion generation requires many denoising steps to generate a video. This slow sequential operation mode is not suitable for video encoding. To design efficient video feature extractors based on video diffusion models, we conduct a detailed investigation of the structure of the latent embeddings spawned by these models, in search for the optimal depth and noise level from where to extract representations with a single forward pass through the model. After a brief overview of diffusion models, we detail the zero-shot and trained light probes used for our analysis, and we conclude this section with a summary discussion of the findings.

\subsection{Background on latent diffusion models} 
Video Latent Diffusion Models (LDMs) like Veo3 operate within a compressed, lower-dimensional representation space. During training, given an input video $x\in\mathbb{R}^{F\times H\times W\times C}$ (containing $F$ frames), an encoder $\textit{E}$ maps the data into a spatiotemporal latent code $z_0=\textit{E}(x)$. A forward process corrupts this latent representation over $T$ steps. At an arbitrary noise level $t\in[0,T]$, the corrupted latent $z_t$ is generated by adding Gaussian noise to $z_0$: $z_t=\sqrt{\bar{\alpha}_t}z_0+\sqrt{1-\bar{\alpha}_t}\epsilon$, where $\epsilon\sim\mathcal{N}(0,I)$ and $\bar{\alpha}_t$ dictates the variance schedule. To invert this process, a backbone network, typically a Diffusion Transformer (DiT) or a U-Net, denoted as $f_\theta$, is trained to predict the injected noise $\epsilon$. The network is conditioned on the corrupted latent $z_t$, the noise step $t$, and contextual information $c$ (e.g., text embeddings). The model is optimised via the standard reweighted variational lower bound: $\mathcal{L}_{LDM}=\mathbb{E}_{z_0,\epsilon,t}\left[\|\epsilon-f_\theta(z_t,t,c)\|_2^2\right]$. Note that flow matching models like Wan 2.2, also included in our study, replace this stochastic forward process with a continuous-time deterministic framework based on Optimal Transport. In this setup, for a continuous time step $t \in [0,1]$, the corrupted latent is a linear interpolation between the data and a base noise distribution: $z_t = (1 - t)z_0 + t\epsilon$. Rather than predicting the injected noise, the network $f_\theta$ is optimised to regress the velocity vector field $f_t = \frac{d z_t}{d t} = \epsilon - z_0$ that transports the data to the noise. The model is trained via the flow matching objective: $\mathcal{L}_{FM} = \mathbb{E}_{z_0, \epsilon, t} \left[ ||f_\theta(z_t, t, c) - (\epsilon - z_0)||_2^2 \right]$.

The backbone $f_\theta$ consists of a sequence of $L$ layers (e.g., transformer blocks in the case of a DiT). We denote the intermediate spatiotemporal activations extracted from the $l$-th layer at noise step $t$ as $h_t^{(l)}$. During inference, the model iteratively denoises pure Gaussian noise $z_T$ to recover an estimated clean latent $\hat{z}_0$. Finally, $\hat{z}_0$ is mapped back to pixel space via a decoder $\textit{D}$, yielding the generated video $\hat{x}=\textit{D}(\hat{z}_0)$.  

\begin{figure}[t]
    \centering
    \includegraphics[width=.95\linewidth]{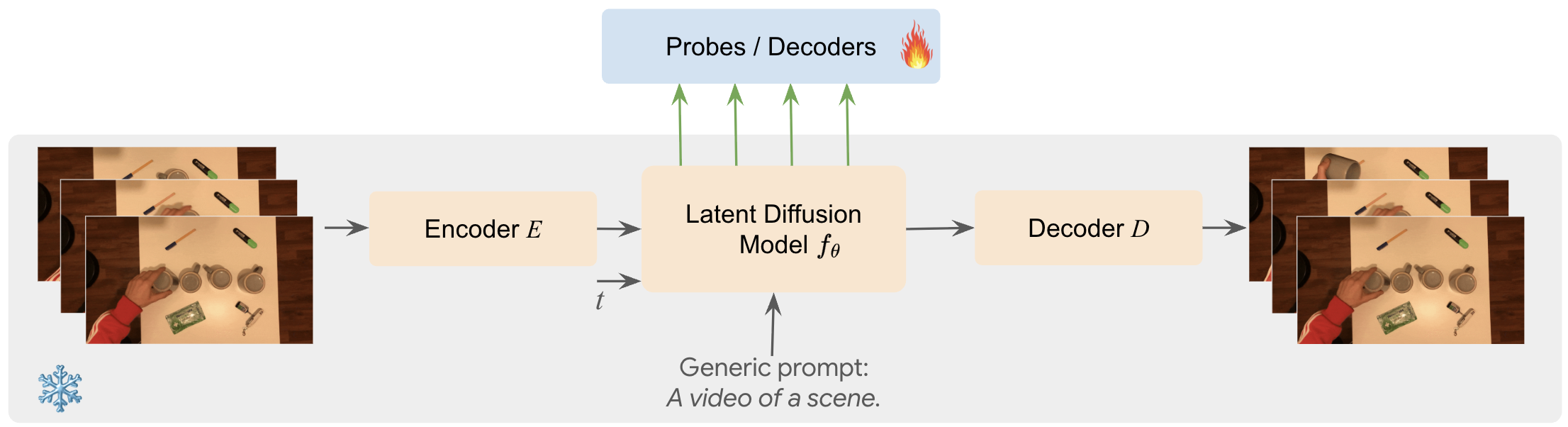}
    \caption{Video generative model repurposed as video encoder.}
    \label{fig:model}
\end{figure}

Figure~\ref{fig:model} illustrates our diffusion-based video encoder \textit{Gen4U}. We condition the diffusion transformer on a generic text embedding: e.g. \textit{A video of a scene}, to not leak the ground truth (action label or caption) for semantic tasks. We have also experimented with conditioning on an empty text string, without significant difference in results. To deal with the inherent stochasticity of the diffusion process, we use a fixed random seed throughout all our experiments to ensure reproducibility. In all our plots, we report results for blocks and noise levels expressed as percentages obtained as a linear fraction of the total depth or noise steps.

\subsection{Zero-shot probes}

\par
\begin{figure}[h]
    \centering
    \includegraphics[width=.97\linewidth]{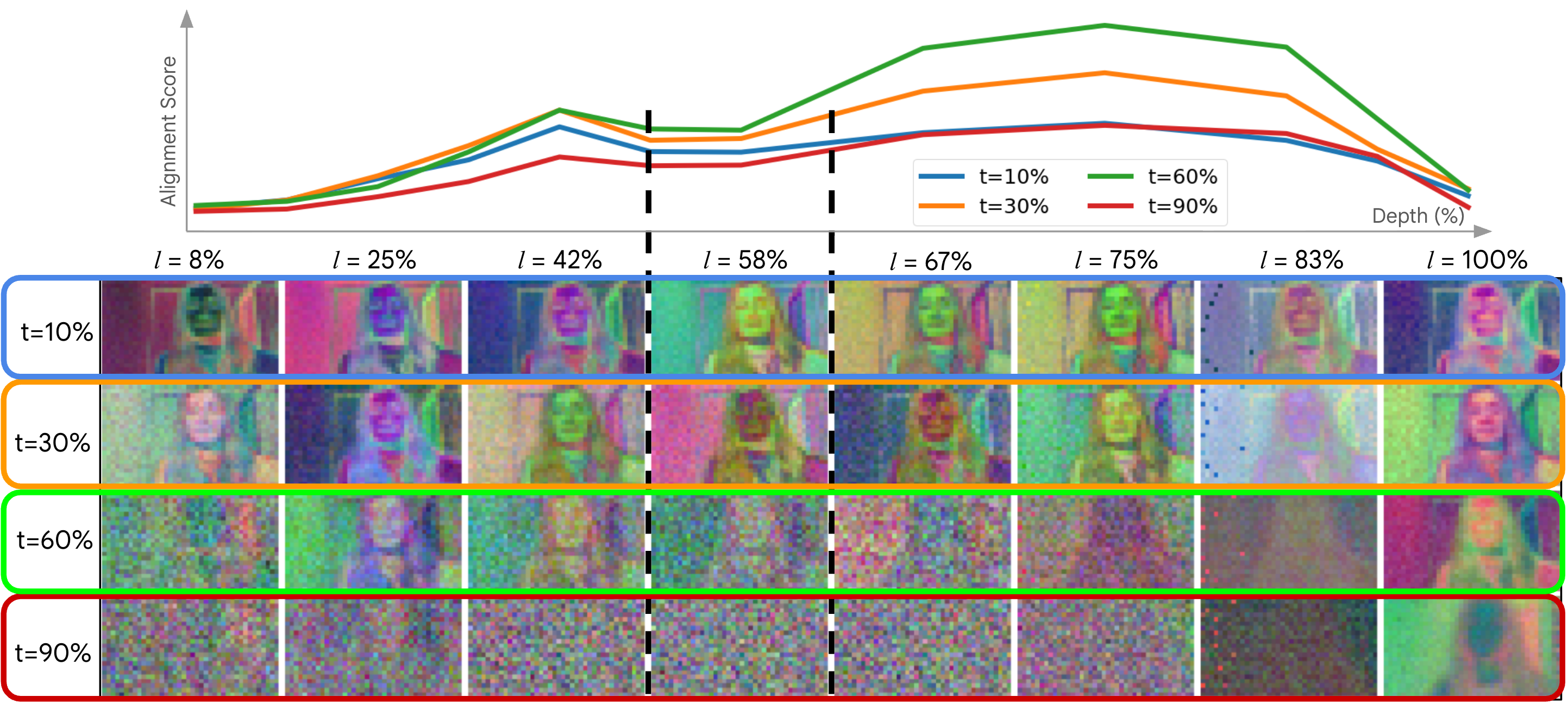}
    \caption{PCA visualisation of activations extracted from Veo3 at different depths and noise levels. For each (depth, noise level) pair, we compute PCA over all spatial tokens aggregated across the dataset. The top three principal components are mapped to RGB channels. Noise level $t=60\%$ (in green) has maximal alignment with language (curves inpainted at the top from Fig.~\ref{fig:align} left) and other visual encoders; see text for details.}
    \label{fig:pca}
\end{figure}

{\textbf{PCA visualisation:}} To qualitatively assess the evolution of the diffusion features along network depth and noise levels, we plot in Figure~\ref{fig:pca} the PCA visualisation of the activations extracted from different depths and noise levels in Veo3. We can observe that at high levels of noise, the features are simple, low-frequency, encoding the general shapes in the scene. At lower noise levels, the features become more complex, encoding higher-frequency, more refined details. This is in line with the intuition that diffusion is spectral autoregression~\citep{dieleman2024diffusion}: at each denoising step, the model learns to produce the features at the next (higher) frequency level.

\par{\textbf{Mutual k-NN alignment:}} To quantitatively assess the structure of the diffusion embedding space, we adopt the zero-shot Mutual $k$-Nearest Neighbours (M$k$NN) alignment metric proposed by \citep{pmlr-v235-huh24a} and extended to video in~\citep{ovsjanikov2026dynamic}. The key idea is to compare the neighbourhood structure induced by the diffusion model's intermediate representations against that of established, independently trained encoders, without requiring any learned mapping between the two spaces.

Concretely, given a dataset of $N$ videos, paired with ground truth captions, we extract a video-level embedding from each video using both the diffusion model and a reference encoder. For the diffusion model, we obtain intermediate activations $h_t^{(l)}$ at a chosen layer $l$ and noise level $t$, and aggregate them spatiotemporally to produce a single vector per video (see appendix~\ref{app:mknn} for details). For the reference encoder, which may operate in a different modality (e.g., a language model) and produce embeddings of a different dimensionality, we similarly obtain one vector per video. In particular, for image models, we follow~\citep{ovsjanikov2026dynamic} and average the individual image features across input frames. We then independently construct the $k$-nearest-neighbour graph in each embedding space and measure the average overlap between the two sets of neighbours. Intuitively, a high overlap indicates that the diffusion model organises videos in a manner consistent with the reference encoder, despite never having been trained with the same objective. More details are included in the appendix~\ref{app:mknn}.
Following prior work~\citep{pmlr-v235-huh24a, ovsjanikov2026dynamic}, we additionally optimise over the choice of intermediate layers in both encoders and report the pair of layers that maximises the alignment score. We use $k{=}10$ and $N{=}1024$ videos.

\begin{figure}[t]
    \centering
    \begin{subfigure}{0.48\textwidth}
        \centering
        \includegraphics[width=\linewidth]{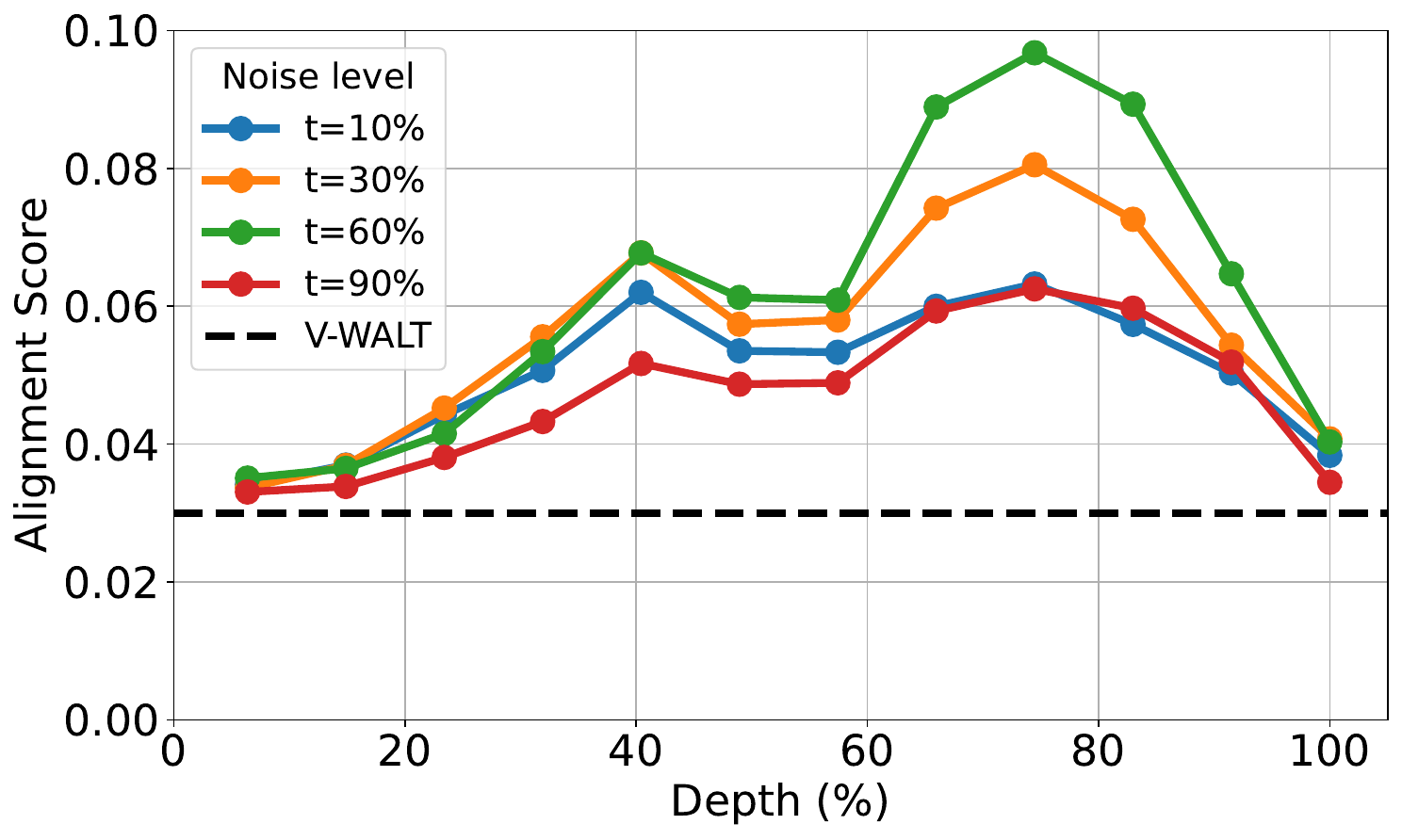}
        \label{fig:align_sub1}
    \end{subfigure}\hfill
    \begin{subfigure}{0.48\textwidth}
        \centering
        \includegraphics[width=\linewidth]{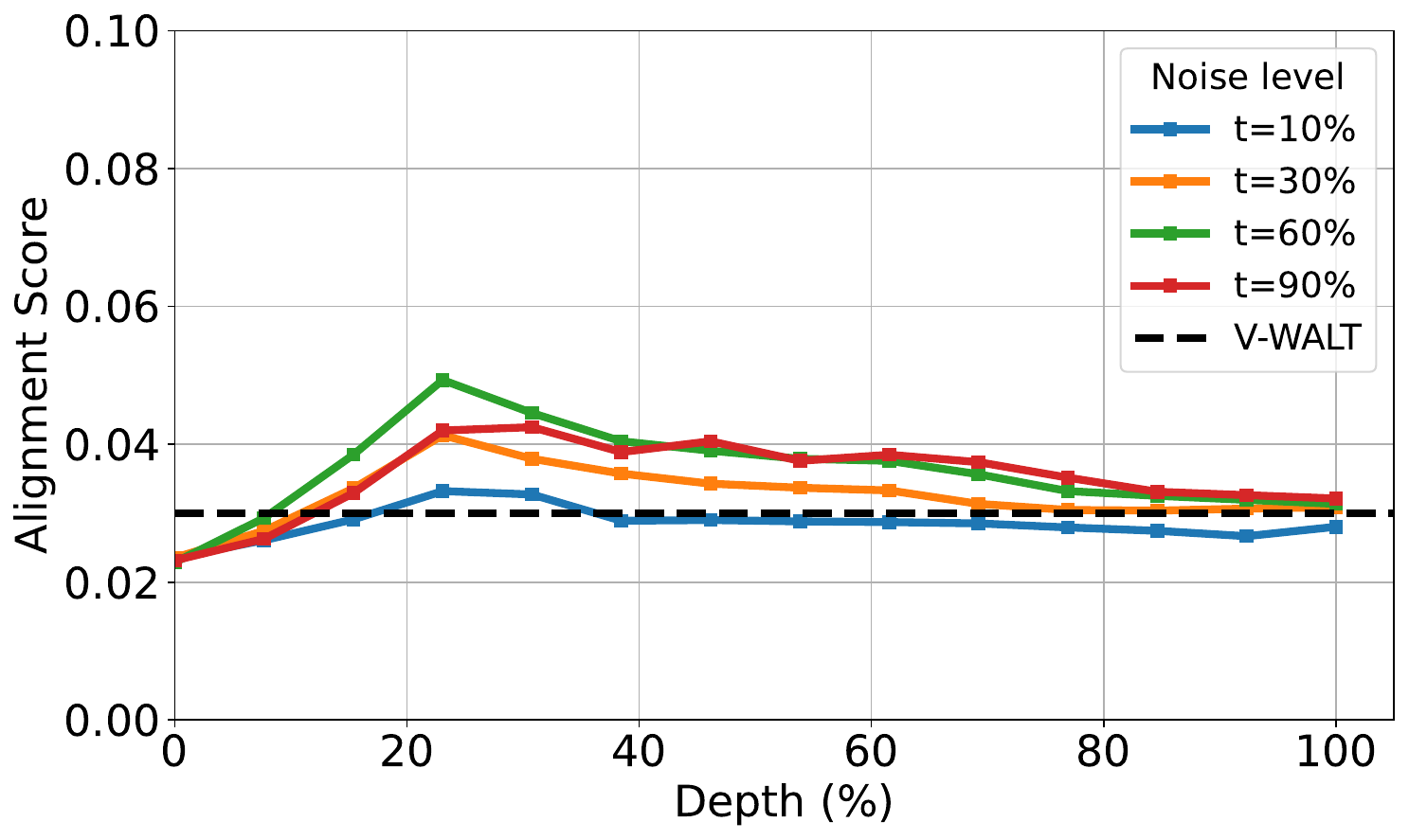}
        \label{fig:align_sub2}
    \end{subfigure}
    
    \caption{Mutual-kNN zero-shot video-text alignment for Veo3 (left) and Wan 2.2 (right) against Gemma-2-9b-it, compared to the best alignment obtained with V-WALT~\citep{Velez_2025_ICCV} for reference. Note that as video generation models become more capable, their internal representations align more strongly with the underlying text captions (not provided to the models). See main text for a detailed discussion.}
    \label{fig:align}
\end{figure}

\begin{figure}[t]
        \centering
    \begin{subfigure}{0.32\textwidth}
        \centering
        \includegraphics[width=\linewidth]{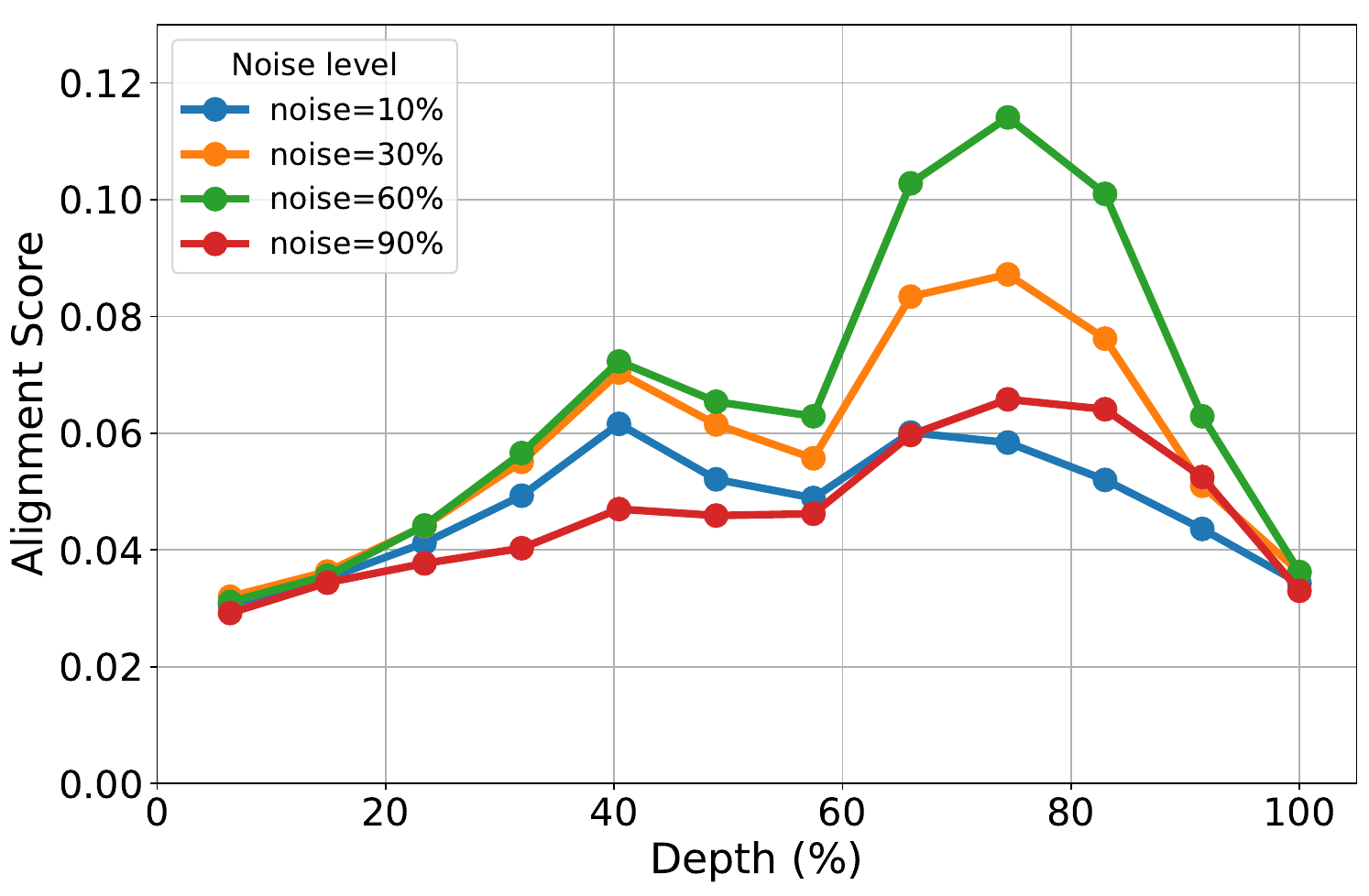}
        \caption*{Veo3 - DINOv2}
        \label{fig:align_non_sem_sub1}
    \end{subfigure}\hfill
    \begin{subfigure}{0.32\textwidth}
        \centering
        \includegraphics[width=\linewidth]{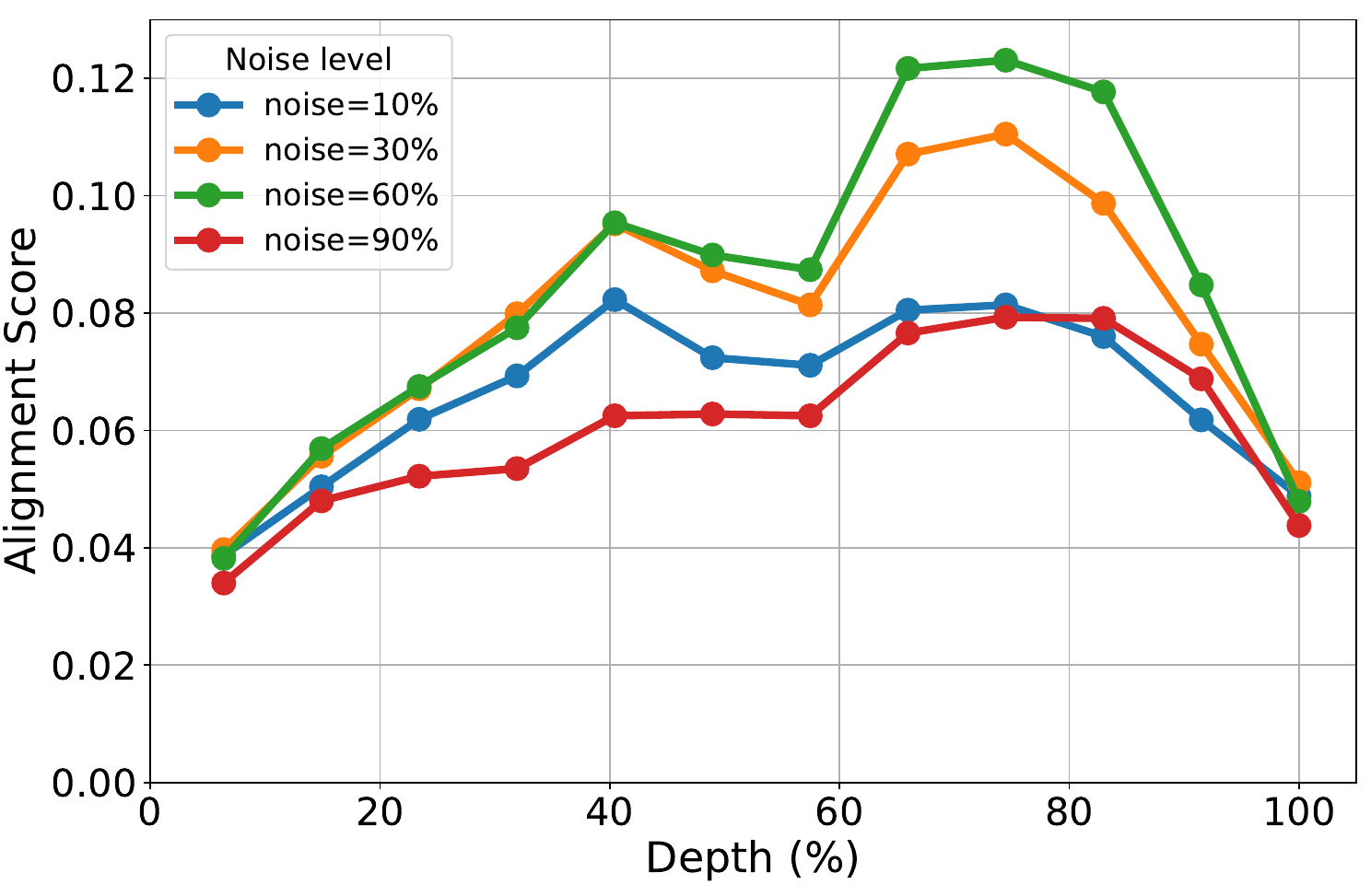}
        \caption*{Veo3 - VideoMAEv2}
        \label{fig:align_non_sem_sub2}
    \end{subfigure} \hfill
    \begin{subfigure}{0.34\textwidth}
        \centering
        \includegraphics[width=\linewidth]{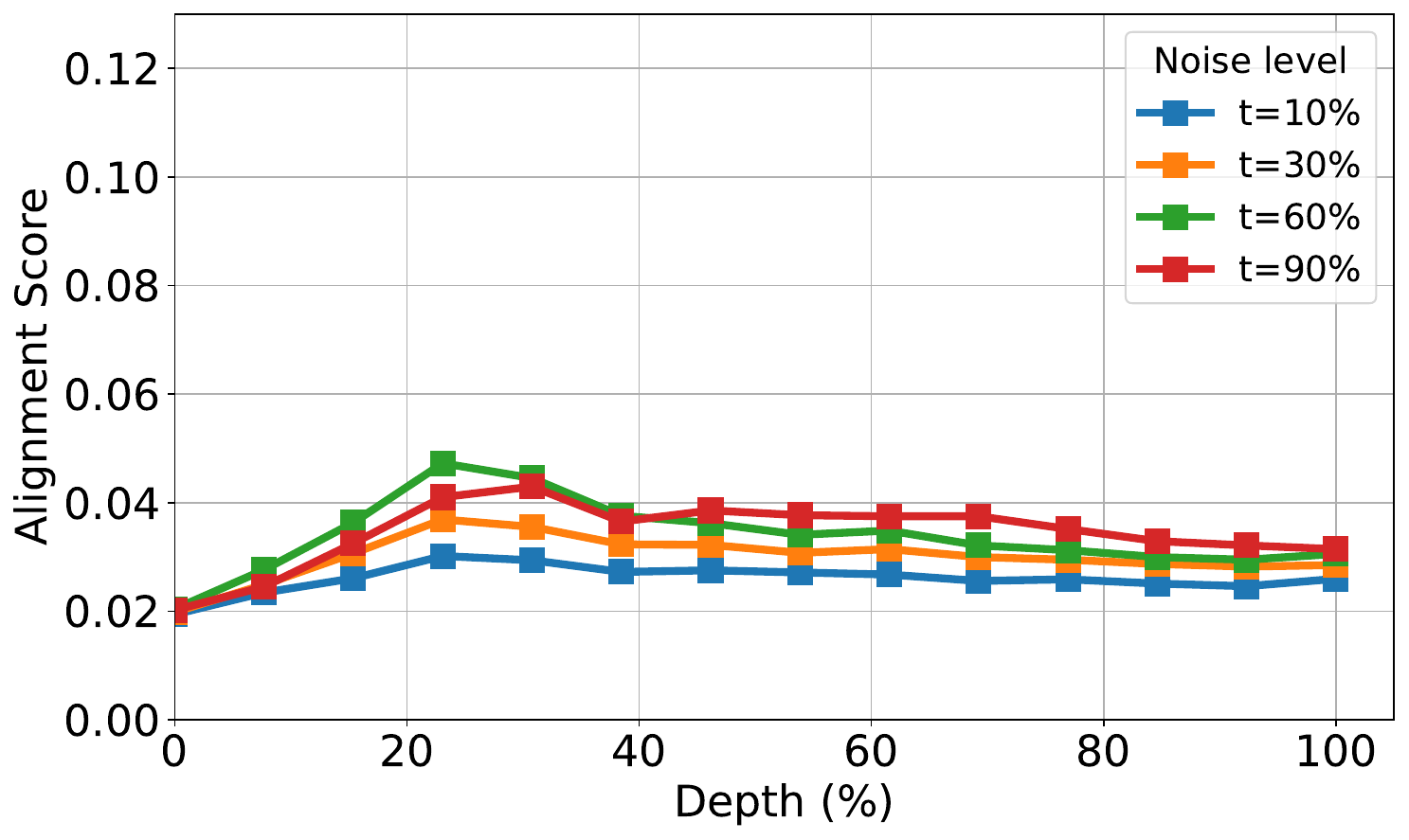}
        \caption*{Wan 2.2 - VideoMAEv2}
        \label{fig:wan_alignment_sub2}
    \end{subfigure} 
    \caption{Alignment between diffusion and discriminative representations using mutual k-NN metric. Remarkably, the intermediate activations of a strong model like Veo3 align not only with underlying semantics (captions) but also with the features of strong image and video models, suggesting their broad applicability.}
    \label{fig:align_non_sem}
\end{figure}

In Figure~\ref{fig:align}, we show the alignment of Veo3 and Wan 2.2 respectively, against the Gemma-2-9b-it text encoder~\citep{gemmateam2024gemma2improvingopen} on the VATEX dataset. This metric has been shown to correlate with downstream performance in semantic and even non-semantic tasks~\citep{ovsjanikov2026dynamic}. In addition, we also extract in Figure~\ref{fig:align_non_sem} the alignment of diffusion representations against DINOv2~\citep{oquab2024dinov}, a powerful image representation model, and VideoMAEv2~\citep{wang2023videomae}, a strong video representation model. Note that the features Veo3, especially when probed at the appropriate depth and noise levels align significantly more strongly with text embeddings of their associated captions. We emphasize that the captions are not given as context during generation. This suggests that a powerful video generative model has the capacity of encoding rich semantic information within its intermediate features. 

\subsection{Linear and attention probes}

\begin{figure}[t]
\centering
\includegraphics[width=0.4\linewidth]{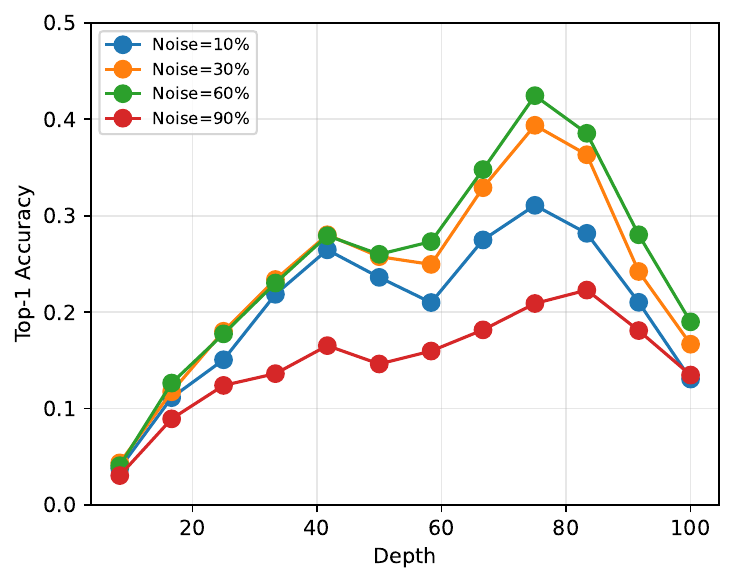}~
\includegraphics[width=0.4\linewidth]{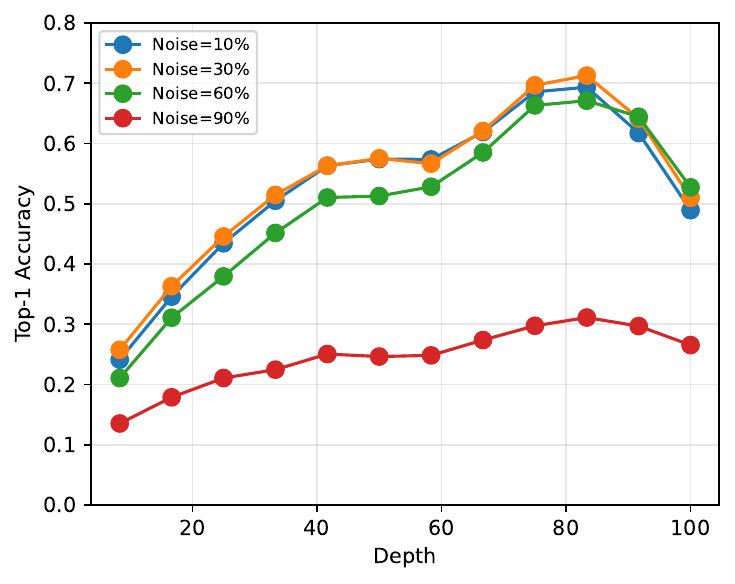}
\caption{Linear (left) and attention (right) probing of Veo3 activations with SSv2 video classification. Observe that both the linear and attention probes peak at around 70-80\% of the model depth and 30-60\% noise level, indicating the optimal spot for extracting representations.}
\label{fig:trained_probes}
\end{figure}

We use linear and attention probes to further assess the quality of diffusion representations. The linear head does global pooling over spatiotemporal dimensions followed by a linear projection. The attention probe uses a cross-attention head, with task-specific queries. We show results for video classification probes in Figure~\ref{fig:trained_probes}.

We also experiment with lightweight decoders that reconstruct the input in RGB space given the representations from different blocks and noise levels, to get a sense of the compression and artefacts present at each stage, but the results are less conclusive; see details and visualisations in appendix~\ref{app:preview}.

\subsection{Discussion}
\label{sec:discussion}

\par\textbf{Feature routing across network depth:} The evolution of diffusion representations follows a dual-axis trajectory along both depth $l$ and noise level $t$. Interestingly, we observe distinct evolution patterns for the two models. The results for Wan 2.2 (Figure~\ref{fig:align} right, Figure~\ref{fig:align_non_sem} right) show a unimodal pattern, aligning with broader diffusion literature (e.g., \citep{luo2023diffusion}). The alignment peaks early, at around 25\% depth, indicating that the model resolves global structure and scene details early before gradually settling finer details. Conversely, Veo3 exhibits a distinct bimodal pattern, a phenomenon not previously reported in the diffusion literature. This mirrors the geometry of representations learnt by large-scale transformers \citep{bimodal}, where the representation manifold expands early on (first peak), contracts significantly in the middle layers to route information, and then expands again toward the end (second peak) before the final output projection. We hypothesise a similar behaviour drives Veo3, supported by PCA visualisations (Figure~\ref{fig:pca}) that show somewhat noisier activations halfway through the network, aligning with the dip in the bimodal pattern. We leave a conclusive characterisation of this behaviour for future work. 

\par\textbf{$\mathbf{60\%}$-noise semantic bottleneck:}
Despite exhibiting divergent topological patterns across depth, both models share a consistent optimal noise level for alignment against both text and visual encoders at $t = 60\%$. This indicates the emergence of a semantic bottleneck for both models at this specific noise level. 

\par\textbf{Evolution of feature complexity, linear vs. attention probes:} Further analysis of Figure~\ref{fig:trained_probes} reveals a slight shift between the optimal extraction points for linear probes ($t \approx 60\%$, $l \approx 70\%$) and attention probes ($t \approx 30\%$, $l \approx 80\%$). While both fall within the second mode of Veo3's bimodal pattern, indicating the sweet spot for semantics, this shift suggests an evolution in feature complexity. At deeper blocks and lower noise levels, linear probes and zero-shot alignment metrics become insufficient possibly because semantic information becomes patch-specific and spatially scattered. Instead, attention probes excel by acting as a dynamic spatiotemporal pooling mechanism, selectively attending to and aggregating these distributed, fine-grained details. This suggests that lower-noise representations do not discard semantic meaning; they simply require more complex probes to decode it. Importantly, the superiority of the attention probe is not merely an artefact of its higher parameter count (6M vs. 1M for the linear probe). If parameter count were the sole driver, the attention head would uniformly dominate across all noise levels. Instead, at extreme noise levels ($t \approx 90\%$, red lines), its performance is severely constrained ($\approx$15–30\%), offering minimal advantage over the linear probe at early-to-mid depths. Finally, all probes experience a sharp performance drop at the deepest layers, indicating that the network discards semantic abstractions at the very end to perform pixel-level signal reconstruction.

\par\textbf{Emergence of general-purpose features:} Overall, these alignment results are remarkable. Prior evaluations of generative video models, such as WALT \citep{ovsjanikov2026dynamic}, demonstrated limited alignment with semantic encoders. In contrast, our findings show that training purely for video generation at scale is sufficient to endow a model with powerful, general-purpose features. While these large-scale diffusion models are extensively trained with text conditioning, they are never explicitly optimised to align their internal states with text representations (unlike contrastive learning). The implicit emergence of this alignment is a significant finding not reported in prior work. Furthermore, both models demonstrate strong alignment with established visual encoders (Figure~\ref{fig:align_non_sem}). Veo3, in particular, exhibits high alignment with DINOv2 and VideoMAEv2—models designed for image and video understanding tasks respectively. This confirms that generative representations possess strong potential for diverse downstream tasks requiring a combination of appearance, motion, and semantic understanding. Finally, the consistently higher alignment scores of Veo3 compared to Wan 2.2 confirm that stronger generative models yield correspondingly richer and more powerful representations. 

\par\textbf{Combination of features:} We investigated whether combining features across multiple stages of the diffusion process could yield complementary information. We include details in Appendix~\ref{sec:appendix_combining}.

\section{Diffusion models as video encoders}
\label{sec:experiments}

We demonstrate the potential of diffusion models to act as general-purpose video encoders by evaluating frozen diffusion representations on diverse semantic and non-semantic downstream tasks. While our latent space analysis in Section~\ref{sec:method} confirmed the presence of optimal extraction points across both open-weight (Wan 2.2) and proprietary (Veo3) models, it also revealed that feature alignment scales directly with generative capability. Therefore, to test our hypothesis and establish the upper bound of the Gen4U framework, we focus our downstream evaluations on the highly scaled Veo3 model. We rely on one-block attention decoders to map diffusion representations to various task embedding spaces, similar to 4DS~\citep{carreira2025scaling4drepresentations}. For captioning, we use a small language model fine-tuned on top of the frozen visual diffusion tokens. Only these lightweight decoders are trained, the generative model stays frozen, preserving its generation capabilities.

\subsection{Semantic understanding}

\begin{table}[t]
    \centering
\resizebox{0.9\textwidth}{!}{
    \begin{tabular}{l|c|c|c}
    \toprule
        Model & Pre-training & Model size (M) & Top-1 accuracy (\%) \\
    \midrule
        VideoMAEv2-g~\citep{wang2023videomae} & MAE & 1,013 & 65.6 \\
        VideoPrism-g~\citep{videoprism} & MAE, Contrastive & 1,113 & 65.4 \\
        4DS-j~\citep{carreira2025scaling4drepresentations} & MAE & 21,495 & 68.2 \\
        InternVideo2~\citep{internvideo2} & MAE, Contrastive, Captioning & 6,000 & 67.7 \\
        V-JEPA-H~\citep{bardes2023vjepa} & Masked feature prediction & 635 & 72.2 \\
        V-Walt~\citep{Velez_2025_ICCV} & Diffusion & 1,900 & 59.7 \\
        \textbf{Gen4U (ours)} & Diffusion & - & 71.3 \\
        \textbf{Gen4U w data augm (ours)} & Diffusion & - & \textbf{72.6} \\
    \bottomrule
    \end{tabular}
}
    \vspace{0.1em}
    \caption{Comparison with state-of-the-art on SSv2 video classification. These methods share the same evaluation protocol: frozen representations are fed into a trained one-block attention read-out.}
    \label{tab:ssv2}
\end{table}

\begin{figure}[t]
    \centering
    \includegraphics[width=0.45\linewidth]{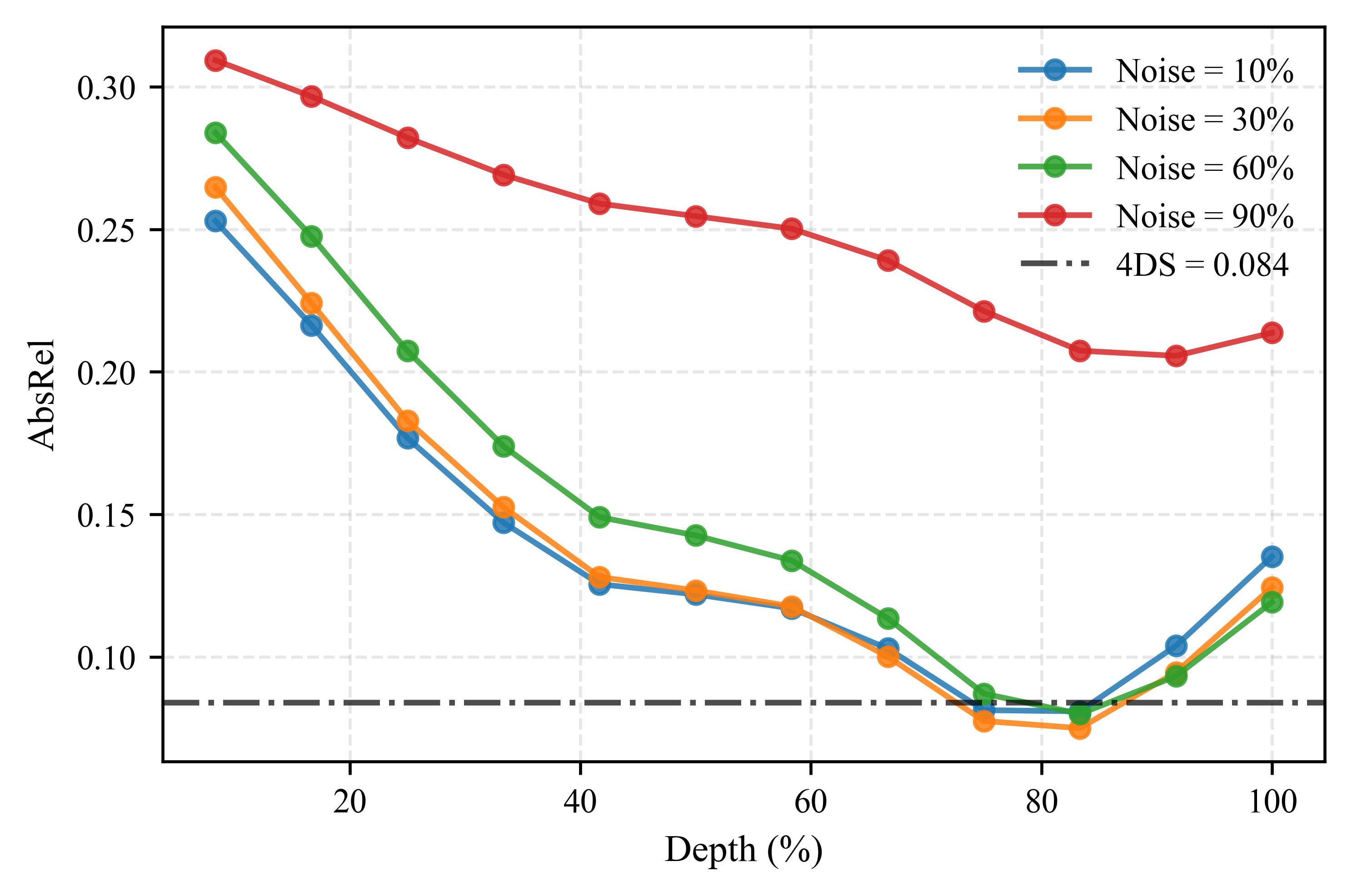}
    \includegraphics[width=0.45\linewidth]{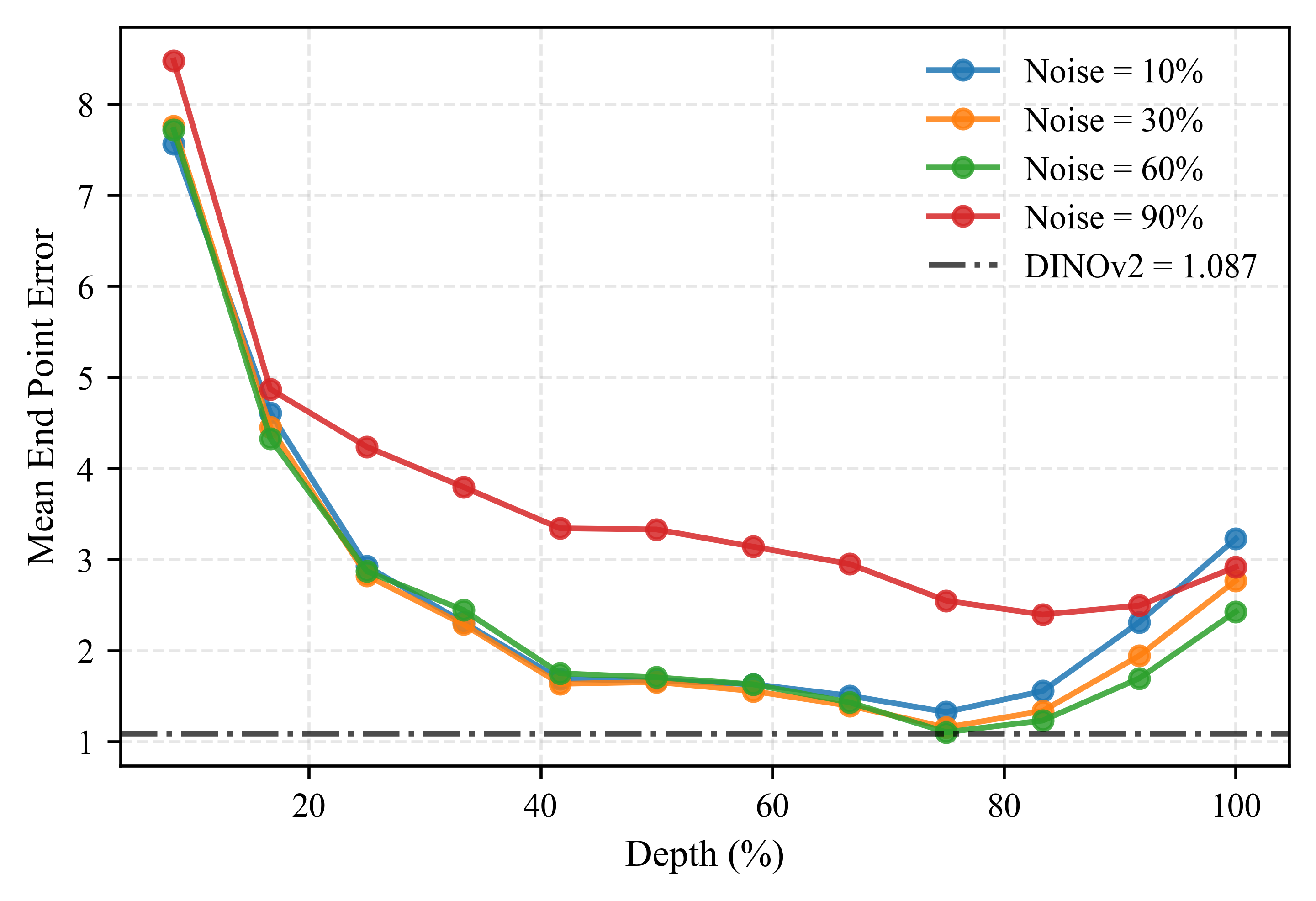}
    \caption{Performance on geometry tasks tasks across blocks and noise levels confirm the sweet spot (depth 75-80\%, noise 30-60\%) identified using alignment and classification probes. Left: Depth estimation AbsRel ($\downarrow$) on ScanNet. Right: Camera pose estimation EPE on ScanNet. The dashed lines mark strong baselines: 4DS~\citep{carreira2025scaling4drepresentations} and DINOv2 respectively.}
    \label{fig:segndepth}
\end{figure}

\par\textbf{Video classification:} We use the challenging SSv2 dataset~\citep{goyal2017something} for this task and follow the protocol in~\citep{carreira2025scaling4drepresentations}, using simple linear and one-block attention decoders on top of frozen diffusion representations. The SSv2 dataset contains 220,847 shorter videos (2-6s long), sampled at 12fps, representing 174 classes. The videos depict actions that differ in finer motion-related details, requiring a deeper temporal understanding, e.g. \textit{pouring something into something} vs \textit{pretending to pour something into something}.

Unlike previous work~\citep{carreira2025scaling4drepresentations}, we do not do data augmentation on the raw videos. Instead, to reduce computation cost, we process only the original videos with Veo3. We then apply data augmentations directly to the intermediate activations: temporal masking (i.e. set to 0) of 16.7\% of the frames, dropout in the attention masking of 40\% and label smoothing of 0.2.

We report performance with and without these augmentations in Table~\ref{tab:ssv2}, compared to strong competitor models. We achieve SOTA performance on this task in this evaluation setup. Note that even without the data augmentation, our setup outperforms all baselines, except V-JEPA.

\par\textbf{Image and video captioning:} In this suite of experiments, we attach the Gemma-2 (2B)~(\citep{gemmateam2024gemma2improvingopen}) text-only large language model as a decoder for image and video
\begin{wrapfigure}[14]{r}{0.5\textwidth}
    \centering
    \begin{subfigure}[b]{0.18\textwidth}
        \centering
        \includegraphics[width=\textwidth]{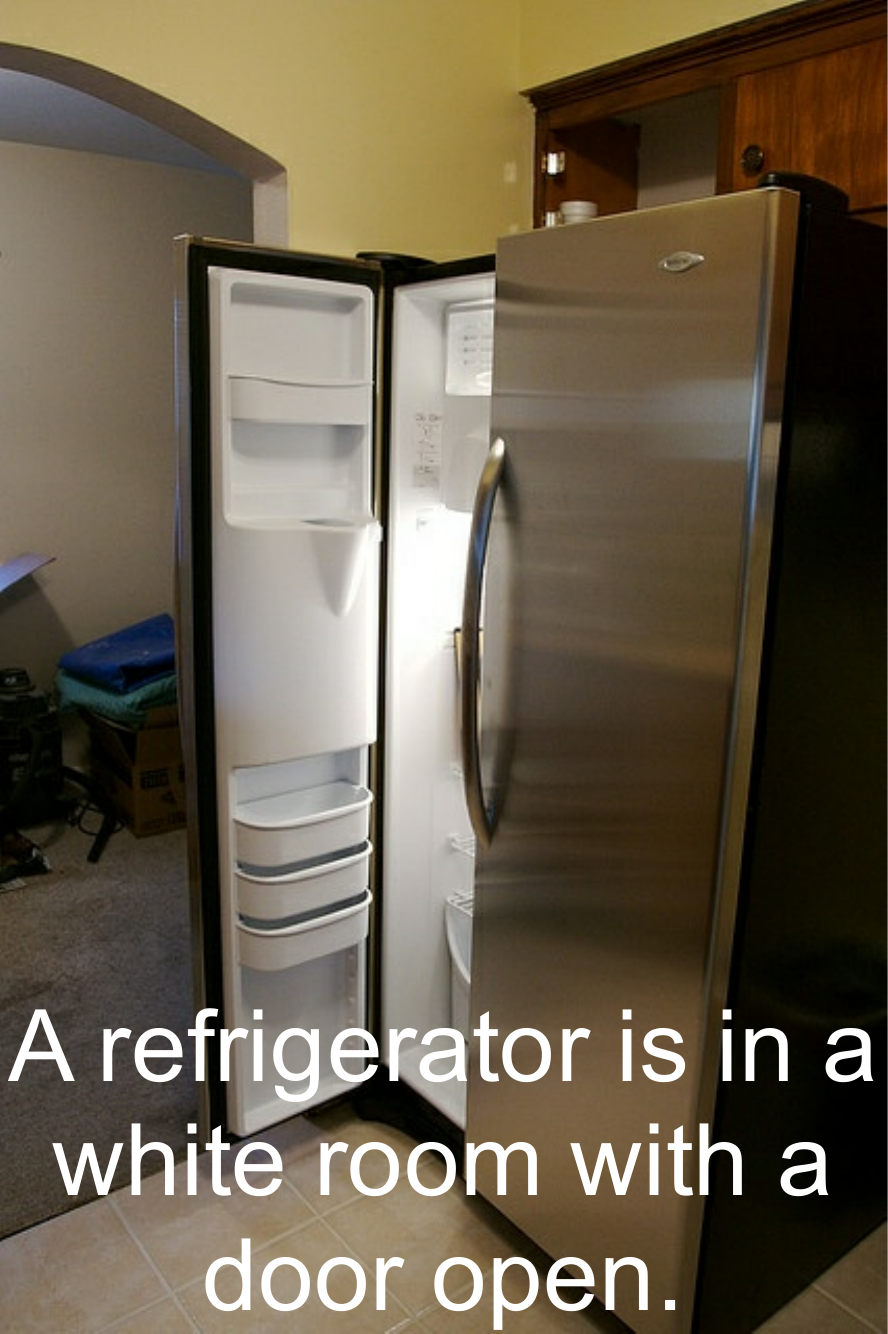}
    \end{subfigure}~
    \begin{subfigure}[b]{0.36\textwidth}
        \centering
        \includegraphics[width=\textwidth]{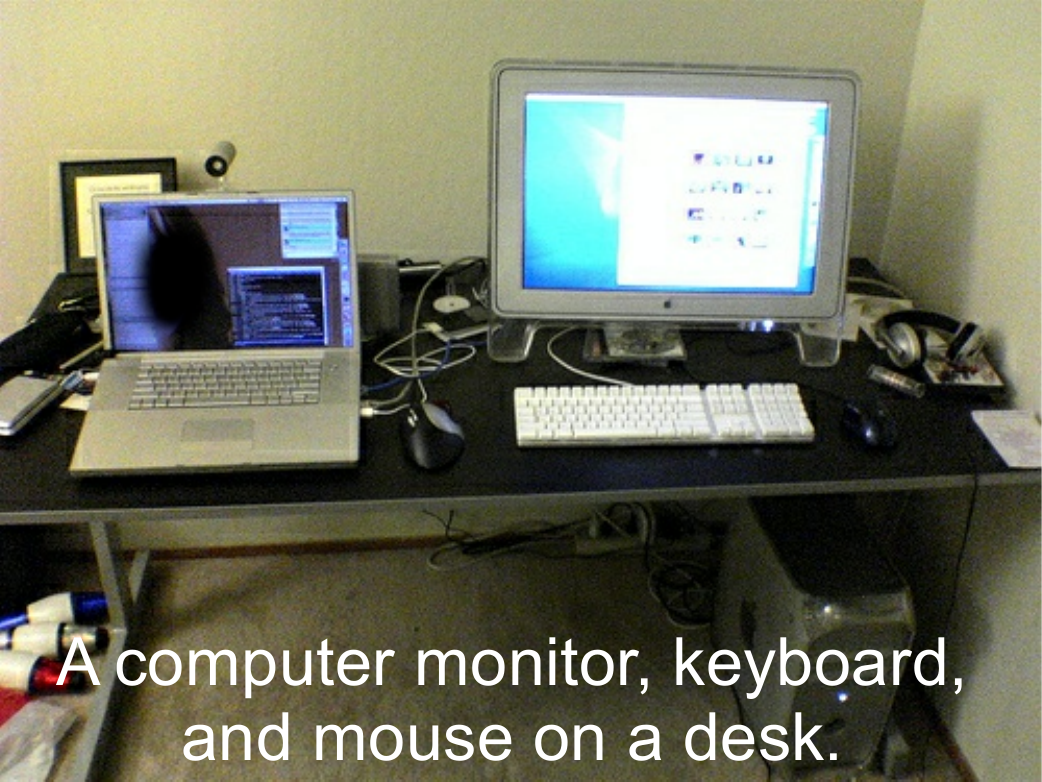}
    \end{subfigure}
    \caption{Captions generated using Gen4U with a Gemma2-2B decoder on the COCO dataset.}
    \label{fig:cap:qualitative}
\end{wrapfigure}
captioning.
Our captioning setup consists of the following two steps: (1.) A cross-attention adapter, inspired by the decoder design of~\citep{sajjadi2022scene}, with three cross-attention blocks, projects the visual representation into $32$ learnable query tokens; and (2.) we prepend these $32$ tokens to the text input before the [BOS] token and process the entire sequence with the LLM using teacher forcing during training and next-token prediction during inference.

We train separate decoders for each of the MS-COCO image captioning dataset (COCO)~(\citep{chen2015microsoft}), the Something-Something V2 video action recognition dataset (SSv2)\footnote{The SSv2 dataset consists of templated class labels like `putting something into something'. We use the labels that incorporate the nouns as well, e.g. `putting an apple into a fruit basket'. We prepend this with `A person is' to form a full sentence: `A person is putting an apple into a fruit basket'.} and the Vatex dataset~(\citep{wang2019vatex}).
We report the CIDEr and BLEU@4 metrics and compare against two baseline models: SigLIP-so400m/14 model from the PaliGemma series~(\citep{beyer2024paligemma,steiner2024paligemma2}) and the SigLIP2-B/16 model~(\citep{tschannen2025siglip}).

\Cref{tab:cap:main} presents the results of Veo3 representations compared to the SigLIP baselines on all three datasets.
Note that our setup does \textbf{not} involve tuning the adapter and LLM decoder on large vision-language data, so it is not directly comparable to SoTA methods like BLIP-2~(\citep{blip2}) and PaLI Gemma~(\citep{beyer2024paligemma}). We train our adapter and LLM only on the train splits of the considered datasets to probe the existence of semantic knowledge decodable through an LLM. 
We observe that video diffusion representations excel on SSv2 but struggle on COCO and Vatex, compared to the SigLIP baselines, which are highly optimised for alignment with text.
Nonetheless, the model produces reasonable captions.
A couple of qualitative examples are shown in Fig.~\ref{fig:cap:qualitative}.

\begin{table}[t]
\centering
\resizebox{1\textwidth}{!}{
\begin{tabular}{l cc cc cc cc}
\toprule
& \multicolumn{2}{c}{\textbf{SSv2}}                     & \multicolumn{2}{c}{\textbf{COCO (Test)}}                                     & \multicolumn{2}{c}{\textbf{Vatex (Test)}}          & \multicolumn{2}{c}{\textbf{Vatex [Frozen LLM] (Test)}} \\
\cmidrule(r){2-3} \cmidrule(l){4-5} \cmidrule(l){6-7} \cmidrule(l){8-9}
\textbf{Model}               & \textbf{CIDEr}           & \textbf{BLEU@4}         & \textbf{CIDEr}           & \textbf{BLEU@4}         & \textbf{CIDEr}          & \textbf{BLEU@4}          & \textbf{CIDEr}          & \textbf{BLEU@4} \\
\midrule
SigLIP-so400m/14             & $204.5 \pm 18.7$         & $34.5 \pm 1.8$          & $\mathbf{118.5 \pm 0.5}$ & $\mathbf{33.2 \pm 0.2}$ & $\mathbf{66.0 \pm 0.8}$ & $\mathbf{34.0 \pm 0.3}$  & $\mathbf{64.5 \pm 0.8}$ & $\mathbf{32.9 \pm 0.2}$ \\
SigLIP2-B/16                 & $198.6 \pm  5.8$         & $33.9 \pm 0.6$          & $114.2 \pm 2.3$          & $32.4 \pm 0.5$          & $58.4 \pm 0.8$          & $31.4 \pm 0.4$           & $56.2 \pm 0.9$          & $30.5 \pm 0.1$ \\
\textbf{Gen4U (ours) @ 30\%} & $\mathbf{289.5 \pm 7.7}$          & $\mathbf{45.5 \pm 0.7}$          & $54.9  \pm 0.8$          & $20.2 \pm 0.2$          & $44.8 \pm 1.5$          & $28.1 \pm 0.5$           & $40.2 \pm 0.4$          & $26.5 \pm 0.2$ \\
\midrule
\textbf{\quad + Noise Aug.}  & $280.4 \pm 16.5$        & $44.8 \pm 1.7$           & $69.3 \pm 1.0$           & $23.5 \pm 0.3$          & $56.7 \pm 0.1$          & $32.3 \pm 0.1$           & $48.5 \pm 0.5$          & $29.6 \pm 0.2$ \\
\textbf{\quad + Noise Aug. + High res.} & -             & -                       & $102.0  \pm 1.3$         & $30.4 \pm 0.4$          & -                       & -                        & -                       & - \\
\bottomrule
\end{tabular}
}
\vspace{0.1em}
\caption{Captioning results across SSv2, COCO, and Vatex Datasets reporting CIDEr $\left( \uparrow \right)$ and BLEU@4 $\left( \uparrow \right)$ mean and standard error across 5 seeds.}
\label{tab:cap:main}
\end{table}

\textit{Vatex [Frozen LLM]:} We freeze the LLM in this experiment and train only the cross-attention adapter on the Vatex train split. Results are reported in the last two columns of Table~\ref{tab:cap:main}. Model performance decreases only slightly relative to the full fine-tuning experiments. Moreover, we significantly outperform VideoPrism which reports a CIDEr of $31.7$ on this dataset. Although VideoPrism does not finetune their adapter on captioning, they pre-train it on a large vision-language corpus.

\textit{Data augmentation using multiple noise injection levels:} As mentioned above, we do not pre-train our adapter and LLM on large datasets. To limit the overfitting caused by small training sets, we augment the data by concatenating the datasets across various noise injection levels $10\%, 30\%, 60\%~\mathrm{and}~90\%$. We find that for COCO and Vatex using the 10\%, 30\% and 60\% noise injection levels is optimal. The model is evaluated on the 30\% level in all cases. The other noise levels purely serve as data augmentation.
For SSv2, we do not see any improvements with data augmentation, likely because the caption space is simpler and the training dataset is much larger than with COCO and Vatex.
The resulting performance is shown in~\cref{tab:cap:main} under \textit{`+ Noise Aug.'}.

\textit{Benefits of high-resolution Veo:} COCO images contain several small objects and the model needs to mention a few of them to score highly on the CIDEr metric. We ablate doubling the input resolution. We pick the optimal combination of noise levels (`+ Noise Aug.') for each resolution on the validation split and report test set numbers in Table~\ref{tab:cap:main} under \textit{`+ Noise Aug. + High res.'}. The high res. model achieves gains $+ 32.7$ CIDEr points. Thus by using multiple noise levels as data augmentation and using higher resolution diffusion representations, one can reduce the gap to discriminative models trained explicitly for the captioning task on large-scale vision-language datasets.

\subsection{Geometry understanding}
\par\textbf{Depth estimation: } We evaluate monocular depth prediction on ScanNet~\citep{dai2017scannet}. Following prior work~\citep{carreira2025scaling4drepresentations}, we report the absolute relative error (AbsRel), computed as $|d^{*} - d|/(d + \epsilon)$ where $d^{*}$ and $d$ denote predicted and ground-truth depth, respectively, and the threshold accuracy $\delta_1$. Target depth values outside the $(0.001, 10)$\,m range are masked out. We use a Dense Prediction Transformer (DPT) readout head~\citep{ranftl2021dpt} with approximately 23M trainable parameters, trained with the scale-invariant logarithmic (SiLog) loss~\citep{eigen2014depth}. The entire diffusion backbone remains frozen.

Figure~\ref{fig:segndepth} (right) presents AbsRel across all blocks and noise levels. The results confirm the sweet spot for extracting representations around (depth 80\%, noise 30\%), identified using zero-shot alignment and classification probes. The best configuration achieves AbsRel of \textbf{0.075} and $\delta_1{=}0.952$, a 10.7\% relative improvement over the frozen-feature baseline of $0.084$ reported in~\citep{carreira2025scaling4drepresentations}. To our knowledge, this is the best result on ScanNet depth with frozen video model features.

\par\textbf{Camera pose estimation:} We are interested in probing the ability to infer the 6DoF relative camera poses from diffusion representations. We use the setup and metric introduced in~\citep{carreira2025scaling4drepresentations}, applied to Scannet dataset. Given diffusion representations for a clip of $F$ frames, we train a one-block attention decoder to predict the relative pose between the first and last frame of the clip, in the form of a $12D$ vector encoding an SE(3) pose transformation ($3\times3$ estimated rotation matrix and a $3\times1$ translation vector). As evaluation metric, we use end-point-error
(EPE), which considers the rotation and translation components jointly; see~\citep{carreira2025scaling4drepresentations} for more details. We obtain 1.10 EPE, which is on par with a strong DINOv2 baseline obtaining 1.08 EPE. The sweep across blocks and noise levels (Figure~\ref{fig:segndepth} left) is consistent with the findings for the other tasks, with performance peaking reliably in the mid-to-late stages of the network ($\sim$75\% depth), at medium noise level (60\%).
\section{Conclusion}
\label{sec:conclusion}

We show that state-of-the-art large-scale video diffusion models can act as competitive video encoders on a diverse set of tasks (video classification, depth estimation, camera pose estimation, video captioning), supporting the first successful attempt to unify video generation and video understanding. Using zero-shot probes and trained lightweight decoders, we investigated the structure of the diffusion latent space and identified the optimal block and noise level to extract representations for understanding tasks. All our results for downstream tasks use representations extracted from a single block and noise level, so we only need to run a single forward pass through the model, being as efficient as large-scale discriminative visual encoders. As future work, we aim to study further the bimodal pattern identified for Veo3 and push the performance to achieve SOTA on a wide range of tasks. 

\noindent \textit{Limitations:} Our experiments are conducted mainly on a proprietary video model, Veo3, with analysis replicated on an open-source model (Wan 2.2), limiting the reproducibility of our study. However, we believe that this is still a useful investigation for the community and we hope it will encourage more work in studying large-scale diffusion models for representation learning. 

\section*{Acknowledgments} We are deeply grateful to Rahul Sukthankar, Howard Zhou, Daniel Zoran, Mehdi S. M. Sajjadi, Forrester Cole, Jo\~ao Carreira, Shiry Ginosar, and Andrew Zisserman for their support and insightful feedback throughout this project.    

\bibliographystyle{plainnat}
\bibliography{biblio}
\newpage
\appendix

\section{Mutual $k$-NN alignment metric}
\label{app:mknn}
We describe the Mutual $k$-Nearest Neighbours (M$k$NN) alignment metric used throughout the paper, following~\citep{pmlr-v235-huh24a, ovsjanikov2026dynamic}.
\paragraph{Notation.}
Consider a dataset of $N$ videos $\{v_i\}_{i=1}^{N}$, each paired with some text $c_i$. We compare the representations of two encoders whose embedding dimensionalities may differ.
\paragraph{Diffusion encoder.}
To use the LDM $f_\theta$ as a video encoder, we run the model in its forward (training) mode at a chosen noise level $t$ and extract the intermediate activations $h_t^{(l)} \in \mathbb{R}^{F' \times H' \times W' \times D}$ from layer $l$ of the backbone, where $F'$, $H'$, $W'$ are the spatiotemporal dimensions of the latent and $D$ is the channel dimension. We aggregate these activations via global average pooling over the spatiotemporal axes to obtain a single embedding vector per video:
\begin{equation}
  \mathbf{x}_i = \mathrm{AvgPool}\!\bigl(h_t^{(l)}(v_i)\bigr) \in \mathbb{R}^{D}.
\end{equation}
We stack all video embeddings into a matrix $X \in \mathbb{R}^{N \times D}$.
\paragraph{Reference encoder.}
A reference encoder $E_{\mathrm{ref}}$ (e.g.\ a text encoder applied to text descriptions, or an independently trained video encoder) maps each input to an embedding of potentially different dimensionality $D'$:
\begin{equation}
  \mathbf{y}_i = E_{\mathrm{ref}}(c_i) \in \mathbb{R}^{D'}, \quad Y \in \mathbb{R}^{N \times D'}.
\end{equation}
When the reference encoder is a pure image model (e.g.\ DINOv2), we average its frame-level features across the temporal dimension, following \citep{ovsjanikov2026dynamic}.
\paragraph{Feature preprocessing.}
Before computing nearest neighbors, we clip each feature matrix to suppress outliers: for a chosen quantile $q$ (we use $q{=}0.95$), we compute $\tau = \mathrm{Quantile}_{q}\!\bigl(|X|\bigr)$ and clip all values to $[-\tau, \tau]$. We then $\ell_2$-normalise each row so that similarities reduce to dot products.
\paragraph{Nearest-neighbour graphs.}
Let $\mathcal{N}_k^X(i)$ denote the set of $k$ nearest neighbours of sample $i$ in the embedding space $X$, computed via dot-product similarity (excluding self-matches). We define an analogous set $\mathcal{N}_k^Y(i)$ for space $Y$. Equivalently, we can encode these as binary indicator matrices $M^X, M^Y \in \{0,1\}^{N \times N}$, where
\begin{equation}
  M^X_{ij} =
  \begin{cases}
    1 & \text{if } j \in \mathcal{N}_k^X(i), \\
    0 & \text{otherwise.}
  \end{cases}
\end{equation}
\paragraph{Alignment score.}
The M$k$NN alignment between $X$ and $Y$ is the mean fraction of shared neighbours:
\begin{equation}
  \mathcal{A}_{\mathrm{M}k\mathrm{NN}}(X, Y) = \frac{1}{kN} \sum_{i=1}^{N} \sum_{j=1}^{N} \bigl(M^X \odot M^Y\bigr)_{ij}
  = \frac{1}{N} \sum_{i=1}^{N} \frac{\bigl|\mathcal{N}_k^X(i) \cap \mathcal{N}_k^Y(i)\bigr|}{k},
  \label{eq:mknn}
\end{equation}
where $\odot$ denotes element-wise multiplication. This score ranges from $0$ (no overlap) to $1$ (perfect agreement) and is invariant to the dimensionality and scale of each space, since it depends only on the rank-order of pairwise similarities.
\paragraph{Layer optimisation.}
Both the diffusion backbone and multi-layer reference encoders expose representations from multiple intermediate layers. Following~\citep{pmlr-v235-huh24a}, we sweep over all pairs of layers $(l, l')$---where $l$ indexes a layer in $f_\theta$ and $l'$ a layer in $E_{\mathrm{ref}}$---and report the pair that maximises $\mathcal{A}_{\mathrm{M}k\mathrm{NN}}$:
\begin{equation}
  (l^\star, l'^\star) = \argmax_{l, l'} \; \mathcal{A}_{\mathrm{M}k\mathrm{NN}}\!\bigl(X^{(l)}, Y^{(l')}\bigr).
\end{equation}
We use $k{=}10$ in all experiments.

\section{Combining Veo3 latents across blocks and noise levels}
\label{sec:appendix_combining}

In this section, we provide details and preliminary results regarding the experiments mentioned in Section \ref{sec:discussion} on combining Veo3 representations across different blocks and noise levels.

\subsection{Experimental setup and architectures}
To investigate the complementarity of Veo3 representations, we extract features from 12 equally spaced blocks across 4 noise levels ($t \in \lbrace 10\%, 30\%, 60\%, 90\% \rbrace$). Features are averaged across space and time. This results in $4 \times 12 = 48$ distinct D-dimensional feature vectors per video. We evaluate four architectures to map these $48$ feature vectors into a single global video representation:

\begin{itemize}
    \item \textbf{Linear adapter}: This computes a simple weighted sum of the 48 feature vectors:
    \[ F_{combined} = \sum_{i=1}^{48} w_i \cdot F_i \]
    where $w \in \mathbb{R}^{48}$ is a vector of learnable scalar weights. This architecture has the fewest parameters (48) and demonstrated the best generalization.
    
    \item \textbf{Shared MLP}: The input features are transposed to $(D, 48)$ and passed through a series of linear layers with ReLU activations shared across the feature dimension:
    \[ \begin{aligned}
    \text{Input}(48, D) &\to \text{Transpose}(D, 48) \to \text{Linear}(48 \to M) \to \text{ReLU} \\
    &\to \text{Linear}(M \to H) \to \text{ReLU} \to \text{Linear}(H \to 1)
    \end{aligned} \]
    The output is then squeezed to produce the final $1 \times D$ vector. We experimented with configurations such as $M=8, H=8$.
    
    \item \textbf{Self-attention}: A learnable CLS token $c \in \mathbb{R}^{D}$ is concatenated with the 48 features to form a $49 \times D$ matrix. All 49 tokens attend to each other via standard self-attention, with queries, keys, and values projected to $d_k=64$. The output corresponding to the CLS token is extracted and projected back to $D$.
    
    \item \textbf{Cross-attention}: A Perceiver-style architecture where a single learnable query token cross-attends to the 48 features (which are first compressed to $d_k=64$). This significantly reduces parameters compared to full self-attention while allowing dynamic weighting.
\end{itemize}

\subsection{Training objectives}
We considered two primary training losses to train these adapters:

\begin{itemize}
    \item \textbf{Cross-entropy loss} (for direct classification on SSv2):
    \[ \mathcal{L}_{CE} = -\frac{1}{B} \sum_{i=1}^{B} \sum_{c=1}^{C} \mathbf{1}[y_i = c] \log \frac{\exp(\hat{y}_{i,c})}{\sum_{c'=1}^{C} \exp(\hat{y}_{i,c'})} \]
    where $B$ is the batch size, $C$ is the number of classes (174 for SSv2), $y_i$ is the true label, and $\hat{y}_{i,c}$ are the logits produced by adding a linear classification head after the adapter.
    
    \item \textbf{Multi-positive InfoNCE loss} (for cross-modal alignment):
    \[ \mathcal{L}_{InfoNCE} = -\frac{1}{B} \sum_{i=1}^{B} \log \frac{\sum_{j \in \mathcal{P}(i)} \exp(\text{sim}(\mathbf{v}_i, \mathbf{v}_j)/\tau)}{\sum_{k=1}^{B} \exp(\text{sim}(\mathbf{v}_i, \mathbf{v}_k)/\tau)} \]
    Here, $\mathbf{v}_i$ is the $L_2$-normalized adapter output for sample $i$, $\text{sim}(\cdot,\cdot)$ denotes cosine similarity, and $\tau$ is a temperature parameter. The set $\mathcal{P}(i)$ contains the indices of the $k$-nearest neighbors in the text feature space (e.g., Gemma embeddings), acting as multiple positives.
\end{itemize}

\subsection{Results}
\label{sec:appendix_extended_results}

\paragraph{Dataset.} We run preliminary experiments using a subset of the SSv2 training set. Specifically, we keep 100 training samples per class, resulting in a training set with 17400 samples. We evaluate on the full SSv2 validation set. The baseline performance of a linear classifier trained on the single best block and noise level using this subset is 17.04\%. This is a modest performance, but it is significantly better than chance, providing the necessary signal for our investigation; see Table~\ref{tab:combine}.

\begin{table}[]
    \centering
    \begin{tabular}{l|c|c|c}
    \toprule
    Adapter & Training data & Training objective & Accuracy \\
    \midrule
      Best single block (baseline) & -- & -- & 17.04 \\
    \midrule
      Linear & VATEX & Text alignment & 21.4 \\
      Linear & SSv2  & Text alignment  & 21.9 \\
      Cross-attention & SSv2 & Classification & \textbf{24.9} \\
    \bottomrule
    \end{tabular}
    \vspace{2mm}
    \caption{SSv2 classification (linear readout head, no augmentation) with Veo3 features. ``Best Single block'' uses the single optimal block and noise level combination under this classification protocol. ``Combined'' uses a learned adapter to fuse features across 12 blocks and 4 noise levels.   Note that training for alignment with text captions (i.e., \textit{without any} SSv2 class supervision) still leads to improvement in downstream classification with a linear adapter. Training a cross-attention based adapter directly for SSv2 classification yields the best result overall. \vspace{-3mm}}
    \label{tab:combine}
\end{table}



Linear adapters consistently outperform non-linear ones in terms of zero-shot generalization across datasets and tasks. For instance, a linear adapter pre-trained purely for text alignment on VATEX generalised effectively to SSv2 out-of-the-box. Conversely, non-linear models—such as the cross-attention and self-attention architectures—were significantly harder to train. In early experiments, they frequently achieved high alignment scores on the training set while generalizing poorly to downstream classification tasks, indicating severe overfitting.

\paragraph{Training for non-linear adapters.} 
To stabilize the training of higher-capacity non-linear adapters, we found that modifying the batch composition was critical. Specifically, increasing the effective batch size and ensuring a robust ratio of anchors, positives, and negatives prevented the cross-attention models from collapsing. Our optimal configuration utilized 16 anchors per batch. For each anchor, we sampled 10 positive matches and 22 negative matches (yielding 32 samples per anchor). This structured batching was the key heuristic that allowed the cross-attention adapters to converge without overfitting to the training distribution.

\paragraph{Peak performance via cross-attention.} 
While linear adapters are robust and easily generalisable, the carefully tuned cross-attention adapter ultimately yielded the highest peak performance when trained directly on the target distribution. When trained specifically for classification on the SSv2 dataset using the aforementioned batching heuristics, the cross-attention adapter achieved a downstream classification accuracy of 24.9\% on this SSv2 subset. This represents a substantial improvement over the 17.04\% baseline (which utilises a single optimal block and noise level) and also outperforms the best linear adapter's peak of 21.9\% on this SSv2 subset. All of these results were obtained using a simple \textit{linear} readout head without any data augmentation.

\paragraph{Correlation across training objectives.}
Throughout our ablation studies across different training objectives (classification, class-vector alignment, and text-embedding alignment), we observed a highly consistent empirical trend: training the adapter for classification utilising a discrete label signal simultaneously improved all proxy metrics for feature quality. 

Most notably, adapters trained using a standard cross-entropy classification loss consistently exhibited \textit{better} zero-shot cross-modal text alignment (e.g., higher mutual k-NN alignment with Gemma text features) than adapters that were explicitly trained to maximize text alignment. Similarly, 1-NN retrieval accuracy also peaked under the classification objective. We hypothesize that the discrete cross-entropy objective provides a cleaner, more stable gradient signal for the adapter to learn the underlying semantic structure of the VEO representations, compared to the noisy gradients inherently present in multi-positive contrastive alignment over text embeddings.

\section{Preview decoding}
\label{app:preview}

We train linear and attention preview decoders on the Something-Something-V2 dataset~\citep{goyal2017something}, i.e. we attempt to decode early (and efficiently) the final RGB output to get a qualitative glimpse into the features encoded at different blocks and noise levels. We use the mean L2 error on RGB values as the training loss.

\textbf{Linear readout head:} The linear head applies a single learned affine projection from the token embedding space to RGB colour. Given input tokens $\mathbf{X} \in \mathbb{R}^{B \times F \times N \times C}$ ($B$ is the batch size, $F$ is the number of frames, $N$ is the number of spatial tokens per frame, and $C$ is the number of channels per token),
the head computes $\hat{\mathbf{Y}}_{\mathrm{lin}} = \mathbf{X} W + b$ with $W \in \mathbb{R}^{C \times 3}$ and $b \in \mathbb{R}^{3}$. The output is reshaped to $(B, F, H, W, 3)$, where $H \cdot W = N$,
recovering the native spatial layout of the Veo3 token grid.

\textbf{Attention readout head:} The attention head uses cross-attention with learnable queries to decode a target-resolution RGB video. A bank of $F \!\cdot\! H \!\cdot\! W$ query vectors is learned, 
each of dimension $d$ < $C$. 
The input tokens are first flattened to $(B, F{\cdot}N, C)$ and linearly projected to keys and values of dimension $d=512$.
A standard multi-head dot-product attention layer with $h{=}8$ heads (head dimension 64) then computes cross-attention from the learned queries to the projected keys and values, yielding an output of dimension $d{=}512$ per query. A final linear projection maps each query output to 3 (RGB) channels, and the result is reshaped to $(B, F, H, W, 3)$.

Both heads are trained with per-pixel L2 loss and evaluated with MSE.
The optimizer is Adam with a linear warmup (1k steps) to a peak learning rate of $10^{-4}$, with gradient accumulation to an effective batch size of 16.

\begin{figure} 
    \centering
    \begin{subfigure}[t]{0.48\textwidth}
        \centering
        \includegraphics[width=\linewidth,trim={0 0 45cm 4cm},clip]{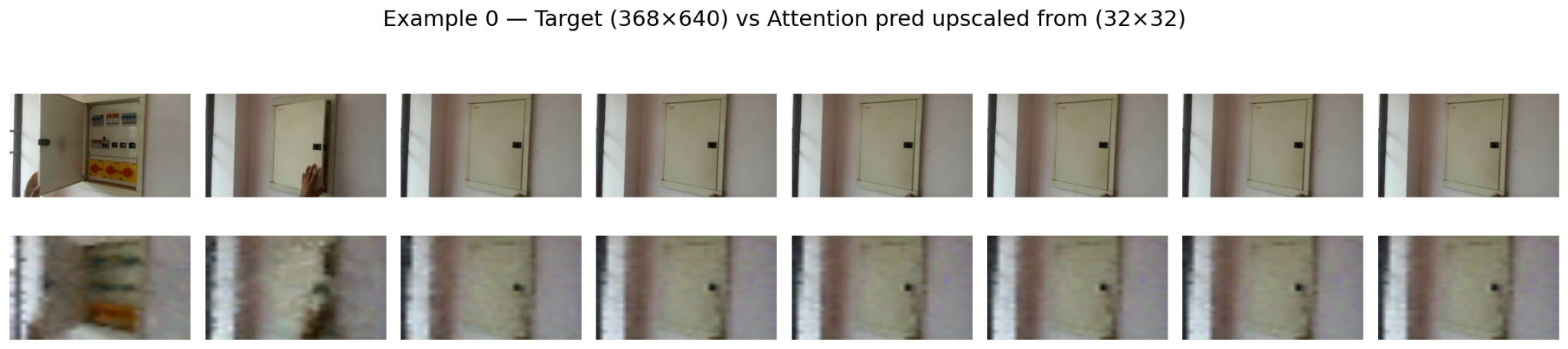}
        \caption{Attention head}
    \end{subfigure}%
    \hfill
    \begin{subfigure}[t]{0.48\linewidth}
        \centering
        \includegraphics[width=\linewidth,trim={0 0 45cm 4cm},clip]{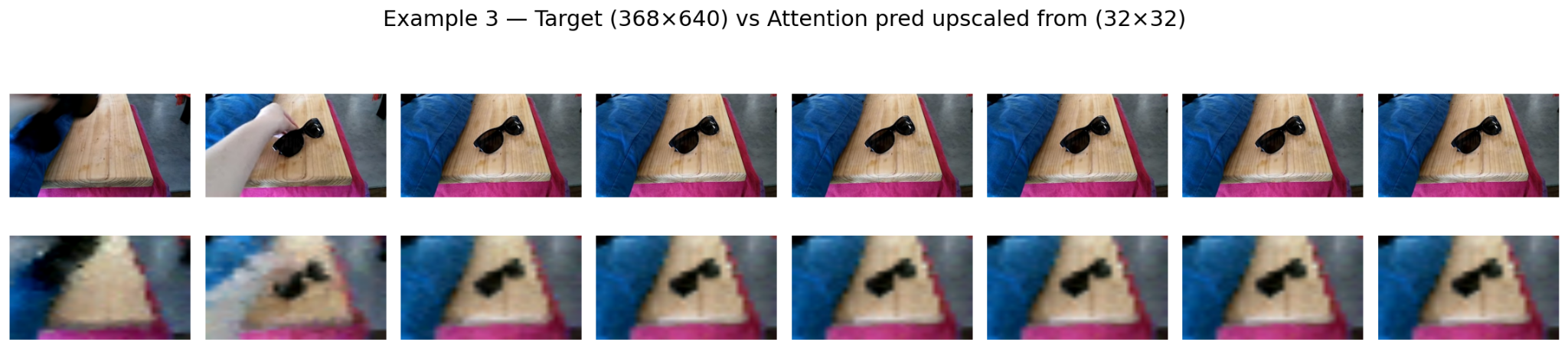}
        \caption{Attention head}
    \end{subfigure}
    
    \vspace{1em}
    
    \begin{subfigure}[t]{0.48\textwidth}
        \centering
        \includegraphics[width=\linewidth,trim={0 0 45cm 4cm},clip]{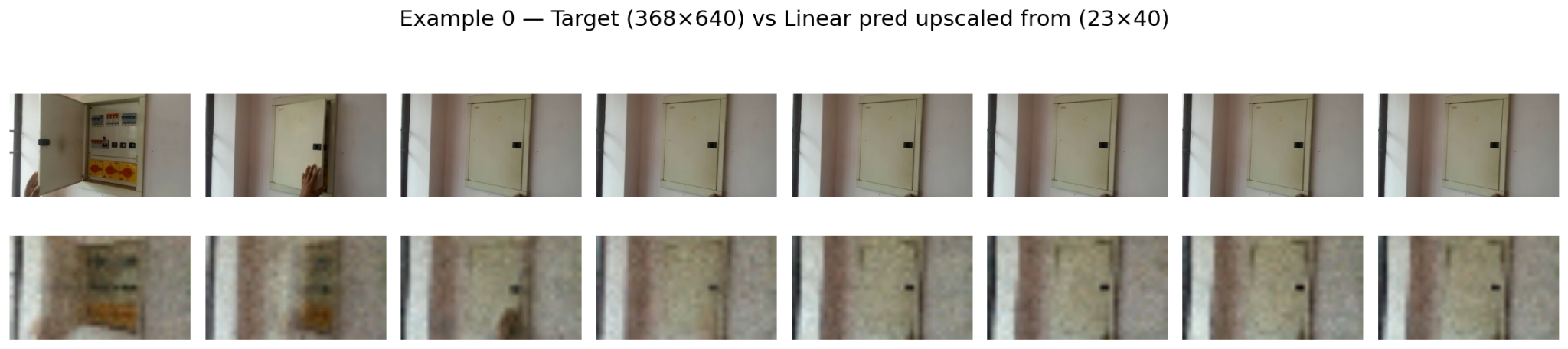}
        \caption{Linear head}
    \end{subfigure}%
    \hfill
    \begin{subfigure}[t]{0.48\linewidth}
        \centering
        \includegraphics[width=\linewidth,trim={0 0 45cm 4cm},clip]{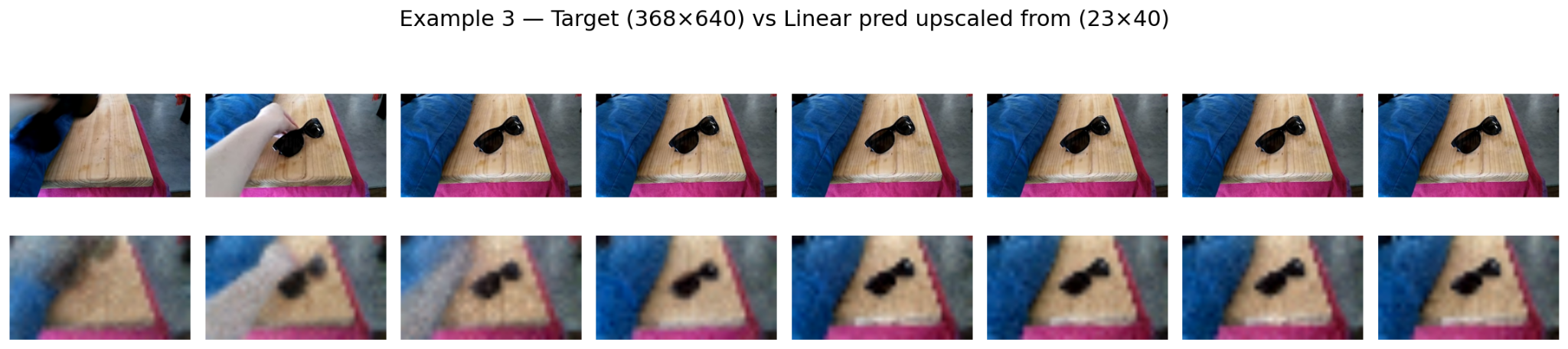}
        \caption{Linear head}
    \end{subfigure}    
    \caption{Qualitative video reconstruction results on two example videos from the Something-Something V2 validation set. In each subfigure, we show  the ground-truth on the top row, and the reconstructions from the linear and single-layer attention head, respectively, on the bottom row. Even with these very simple readout heads, the reconstructions convey well the main elements in the scene.}
    \label{fig:preview}
\end{figure}

\paragraph{Results.} Figure~\ref{fig:preview} presents frames from the original video and their previews decoded from the outputs at depth 75\%, evaluated at noise level 60\%. The linear decoder presents frames that are sharper, but temporally misaligned with the ground truth, as compared to the attention head.

\begin{figure}
    \centering
    \includegraphics[width=\linewidth]{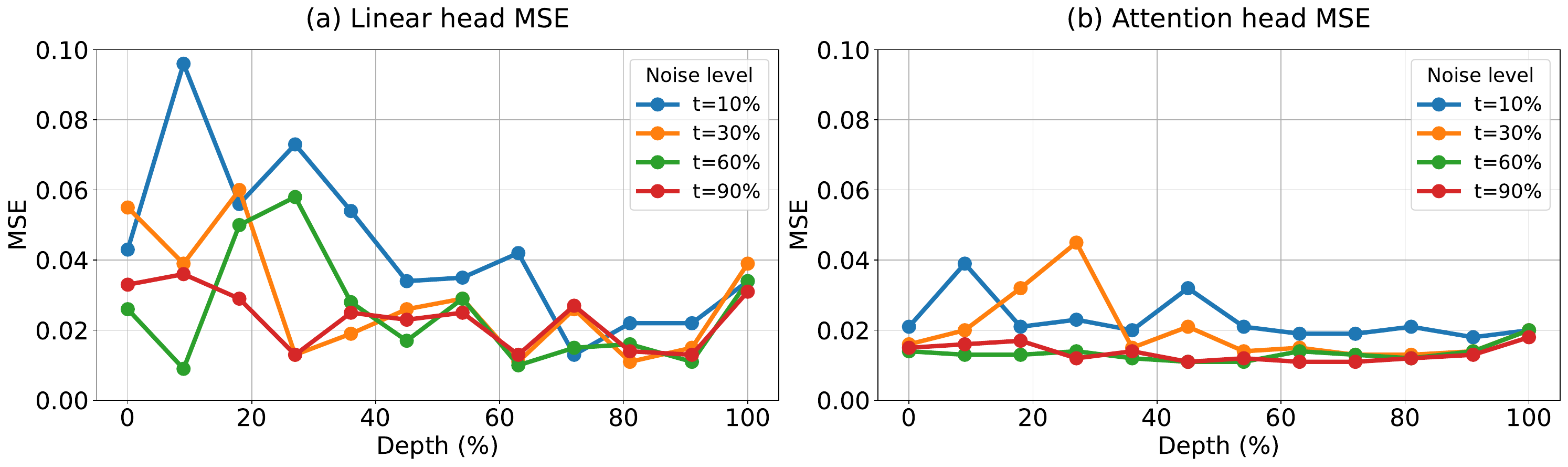}
    \caption{Evaluating RGB readout heads attached at varying network depths and noise levels. Left: linear probe. Right: attention probe.}
    \label{fig:draft_mode_sweep}
\end{figure}

Figure~\ref{fig:draft_mode_sweep} shows MSE reconstruction metrics for attaching the preview head at various network depths, and training it on various levels of injected noise. Unlike the zero-shot alignment scores or the classification probes, regressing the denoised video's pixel values from various heads does not show a clear pattern with an optimal sweet spot for extracting features. The linear head does show generally lower error at higher network depths, reflecting that deeper layers are closer to the final latents to be decoded into RGB values. On average, the attention head outperforms the linear head by having better temporal alignment with the original video.

\section{Image and video captioning details}
\label{app:captioning}
The cross-attention adapter alternates cross-attention layers from learned queries into input vision tokens with self-attention layers across query tokens.
The adapter is designed to output $\mathbf{R}^{2304}$ features matching Gemma2-2B's input embedding size.
Gemma2 is provided an attention mask that is bi-directional on the vision tokens and causal on the text tokens.
The loss is applied to the text tokens \textbf{after} the [BOS] token. During inference we use greedy sampling and discard all tokens after the first [EOS] token is sampled. A maximum sampling sequence length of $64$ tokens is used for the COCO and SSv2 datasets and $96$ tokens are sampled for the Vatex dataset. This was adjusted for Vatex to accommodate longer captions in the dataset.

For SigLIP baselines, input frames are sampled for each video such that the total token count roughly aligns with the number of tokens output by Veo3. The SigLIP-so400m/14 model is from the PaliGemma series~(\citep{beyer2024paligemma,steiner2024paligemma2}) which operates at $896\times896$ resolution with pre-pooling giving us 256 tokens per frame, and the SigLIP2-B/16 model~(\citep{tschannen2025siglip}) operates at $512\times512$ resolution without pre-pooling and outputs 1024 tokens per frame.

Learning rate follows the cosine decay schedule with linear warmup over $1k$ iterations. The peak learning rate is $5\times10^5$.
The Gemma2 weights are initially frozen and gradually thawed with a linear schedule going from $0$ to $0.01$ over $10k$ iterations. It plateaus after $10k$ steps so that the LLM is still updated very slowly to avoid catastrophic forgetting. We use the ADAM optimizer~\citep{kingma2014adam} with batch size of $64$ for all experiments in this subsection.

Models are trained for $32k$ iterations in total. In cases where overfitting was observed, early stopping at $12k$ or $15k$ iterations was applied. This was especially useful on the SSv2 and Vatex datasets when using a single noise injection level as training data (no `Noise Aug.').

The MS-COCO experiments use the Karpathy splits for training, validation and test with 5000 test images~\citep{karpathy2015deep}.

\section{Asset licensing and terms of use}
\label{app:license}
The open-weight models utilised in this study (Wan 2.2, DINOv2) are distributed under the permissive Apache 2.0 license, while Gemma-2 is used in accordance with the Gemma Terms of Use; all three permit broad research and commercial applications. The evaluation datasets (Something-Something V2, COCO, ScanNet, and Vatex) are utilised strictly for non-commercial academic research and benchmarking, in full compliance with their respective institutional Terms of Use and Creative Commons distributions.                                          


\end{document}